\documentclass[10pt,twocolumn,letterpaper]{article}

\usepackage{cvpr}
\usepackage{times}
\usepackage{epsfig}
\usepackage{graphicx}
\usepackage{amsmath}
\usepackage{amssymb}
\usepackage{mathrsfs}
\usepackage{graphicx}
\usepackage{algorithm}
\usepackage{algorithmic}
\usepackage{subfigure}
\usepackage{listings}
\usepackage{booktabs}
\usepackage{appendix}
\usepackage{multirow}
\usepackage{color}
\usepackage[affil-it]{authblk}
\usepackage{flushend,cuted}
\newtheorem{theorem}{Theorem}

\usepackage{mathrsfs}
\makeatletter

\newcommand{\Rmnum}[1]{\expandafter\@slowromancap\romannumeral #1@}
\makeatother

\usepackage[pagebackref=true,breaklinks=true,letterpaper=true,colorlinks,bookmarks=false]{hyperref}

\def\cvprPaperID{1286} % *** Enter the CVPR Paper ID here

\cvprfinalcopy
\begin{document}
%\vspace{-4mm}
%%%%%%%%% TITLE
\title{Learning Adaptive Loss for Robust Learning with Noisy Labels  \vspace{-3mm}} 
%% omit the following lines up until the closing ``}''.
%% Additional authors and addresses can be added with ``\and'',
%% just like the second author.
%% To save space, use either the email address or home page, not both
%\and
\author{Jun Shu~\textsuperscript{1}, Qian Zhao~\textsuperscript{1}, Keyu Chen~\textsuperscript{1}, Zongben Xu~\textsuperscript{1}, Deyu Meng~\textsuperscript{2,1}~\thanks{Deyu Meng is the corresponding author:dymeng@mail.xjtu.edu.cn.} \\
	1. Xi'an Jiaotong University \ \ \ 
	2. The Macau University of Science and Technology\\}

%\author{Sanping Zhou~\textsuperscript{1}, Fei Wang~\textsuperscript{2}, Zeyi Huang~\textsuperscript{3}, Jinjun Wang~\textsuperscript{1}~\thanks{Jinjun Wang is the corresponding author.}\\
%	1. The Institute of Artificial Intelligence and Robotic, Xi'an Jiaotong University\\
%	2. School of Computer Science and Technology, Xi'an Jiaotong University\\
%	3. Robotics Institute, Carnegie Mellon University\\
%}

%\author[1]{Jun Shu}
%\author[1]{Qian Zhao}
%\author[1]{Keyu Chen}
%\author[1]{Zongben Xu}
%\author[2,1]{Deyu Meng*}
%\affil[1]{Xi'an Jiaotong University}  
%\affil[2]{The Macau University of Science and Technology \authorcr *Corresponding author:dymeng@mail.xjtu.edu.cn}

\maketitle
%\thispagestyle{empty}
%\vspace{-4mm}
%%%%%%%%% ABSTRACT
\begin{abstract}
Robust loss minimization is an important strategy for handling robust learning issue on noisy labels.
Current robust loss functions, however, inevitably involve hyperparameter(s) to be tuned, manually or heuristically through cross validation, which makes them fairly hard to be generally applied in practice. Besides, the non-convexity brought by the loss as well as the complicated network architecture makes it easily trapped into an unexpected solution with poor generalization capability. To address above issues, we propose a meta-learning method capable of adaptively learning hyperparameter in robust loss functions. Specifically, through mutual amelioration between robust loss hyperparameter and network parameters in our method, both of them can be simultaneously finely learned and coordinated to attain solutions with good generalization capability.
Four kinds of SOTA robust loss functions are attempted to be integrated into our algorithm, and comprehensive experiments substantiate the general availability and effectiveness of the proposed method in both its accuracy and generalization performance, as compared with conventional hyperparameter tuning strategy, even with carefully tuned hyperparameters.
\end{abstract}
\vspace{-4mm}
%%%%%%%%% BODY TEXT
\section{Introduction} \label{intoduction}
DNNs have recently obtained remarkable performance on various applications \cite{he2016deep,krizhevsky2012imagenet}. Its effective training, however, often requires to pre-collect large scale finely annotated samples. When the training dataset contains certain amount of noisy (incorrect) labels, the overfitting issue tends to easily occur, naturally leading to their poor performance in generalization \cite{zhang2016understanding}. In fact, such biased training data are commonly encountered in practice, since the data are generally collected by coarse annotation sources, like crowdsourcing systems \cite{bi2014learning} or search engines \cite{liang2016learning,zhuang2017attend}. Such robust deep learning issue is thus critical in machine learning and computer vision.

\begin{figure}[t]
	\centering
	%\vspace{-1.3cm}
	\subfigcapskip=-1.5mm
	\subfigure[Adative robust loss in PolySoft]{
		\label{fig1a} %% label for first subfigure
		\includegraphics[width=0.22\textwidth]{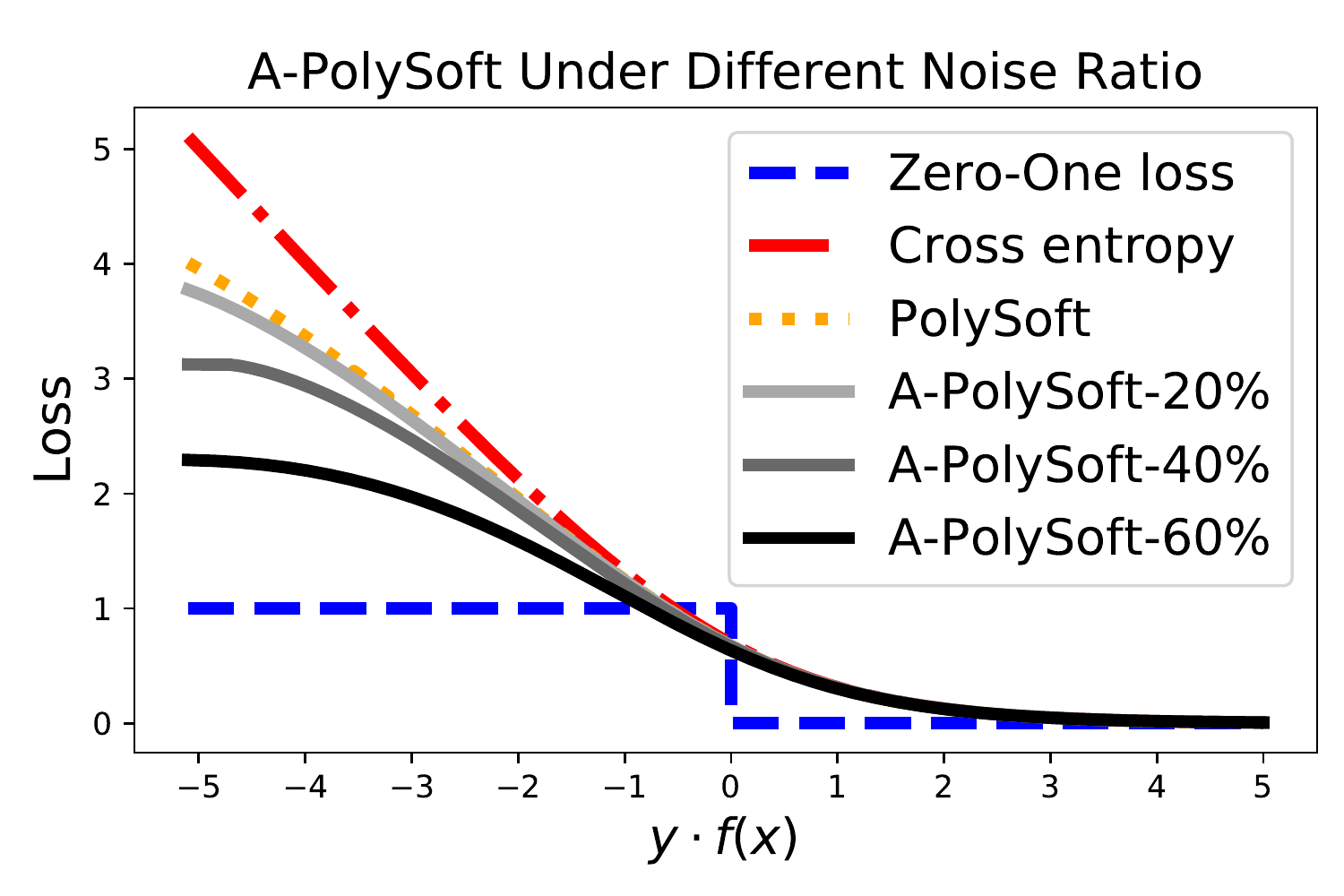}} \ \ \
	\subfigure[Adative robust loss in GCE]{
		\label{fig1b} %% label for first subfigure
		\includegraphics[width=0.22\textwidth]{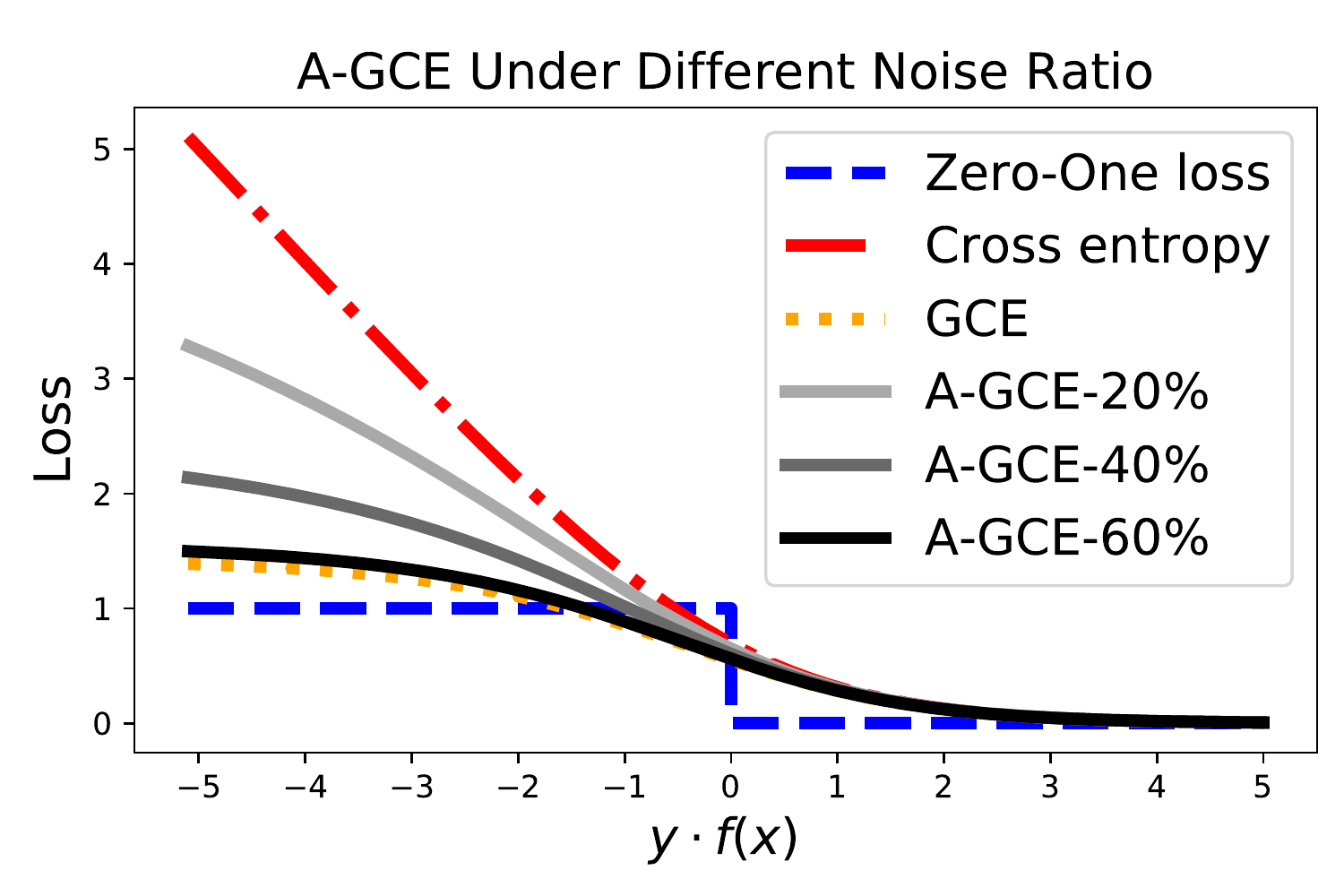}} \\ \vspace{-4mm}
	\subfigure[Adative robust loss in SL]{
		\label{fig1c} %% label for first subfigure
		\includegraphics[width=0.22\textwidth]{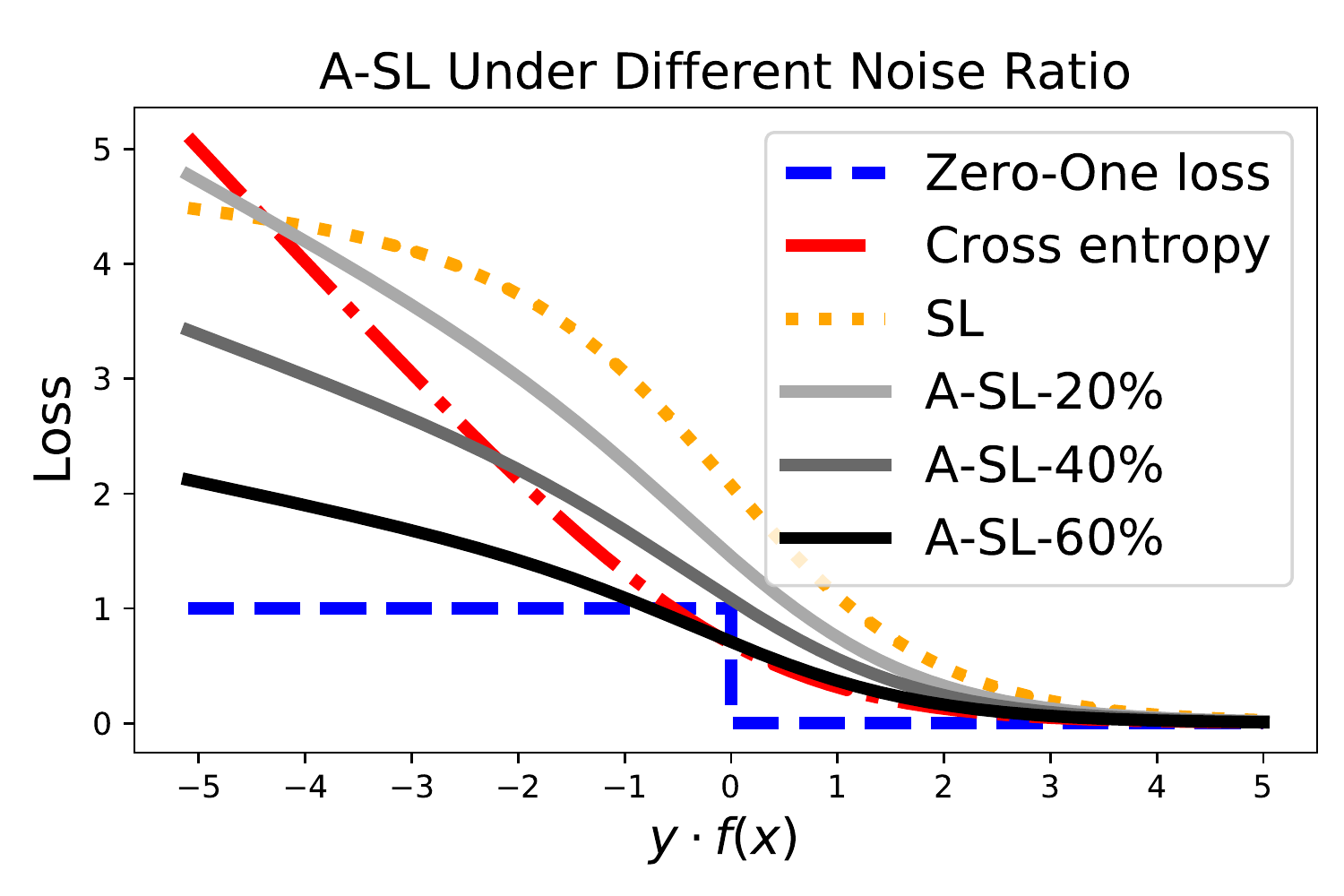}} \ \ \
	 \vspace{-0.2cm}
	\subfigure[Adative robust loss in Bi-Tem]{
		\label{fig1d} %% label for first subfigure
		\includegraphics[width=0.22\textwidth]{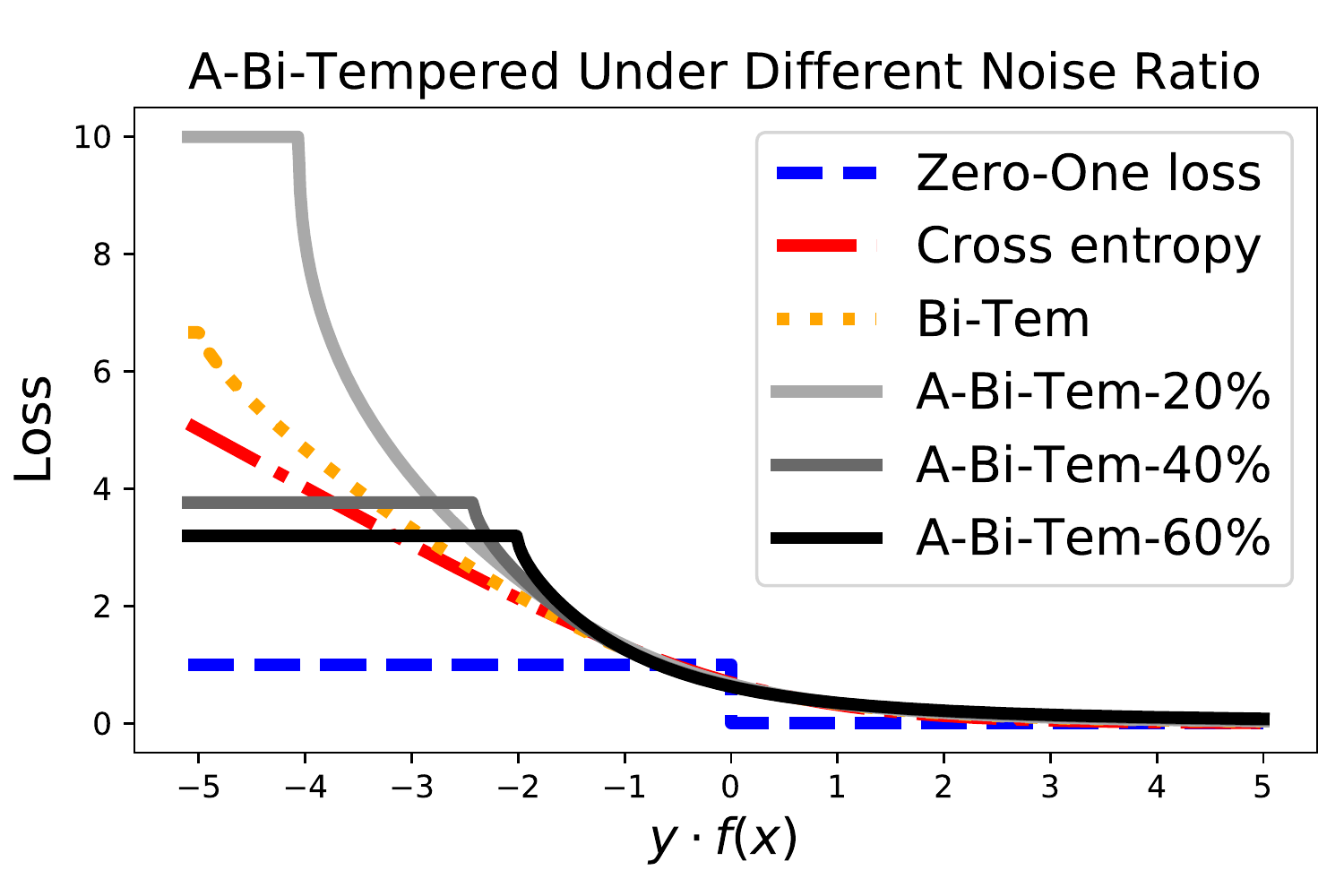}}  \vspace{1mm}
	\caption{Comparison of different loss functions. In each figure, the 0-1 loss, CE loss, original robust loss, and those learned by our method under three different noise rates on CIFAR-10 are shown. The robust losses included in (a)-(d) are PolySoft \cite{gong2018decomposition}, GCE \cite{zhang2018generalized}, SL \cite{wang2019symmetric} and Bi-Tempered \cite{amid2019robust}, respectively.}\label{fig1}
	\vspace{-5mm}
\end{figure}

One of the most classical methods for handling this issue is to employ robust losses, that are not unduly affected by noisy labels, to replace the conventional noise-insensitive ones to guide the training process \cite{manwani2013noise}. For example, as compared with the commonly used cross entropy (CE) loss, the mean absolute error (MAE), as well as the simplest 0-1 loss for classification, can be more robust against noisy labels \cite{ghosh2017robust} due to their evident suppression to large loss values (as clearly depicted in Fig.\ref{fig1}), and thus inclines to reduce the negative influence brought by the outlier samples with evidently corrupted labels. Beyond other robust learning techniques for defending noisy labels, like sample reweighting \cite{kumar2010self,chang2017active,wang2017robust,jiang2018mentornet,ren2018learning,shu2019meta}, loss correction \cite{goldberger2016training,sukhbaatar2014training,patrini2017making,hendrycks2018using}, and label correction \cite{lee2018cleannet,li2017learning,veit2017learning,tanaka2018joint}, such robust-loss-designing methodology is superior in its concise implementation scheme and solid theoretical basis of robust statistics and generalization theory \cite{huber2011robust,liu2014robust,masnadi2009design,patrini2017making}

Along this research line, besides the aforementioned MAE and 0-1 loss, various forms of robust losses have been designed against such robust learning issue on noisy labels. For example, 0-1 loss is verified to be robust for binary classification \cite{manwani2013noise,ghosh2015making}. However, since 0-1 loss is not continuous and the corresponding learning algorithm is hardly to be efficiently and accurately executed, many surrogate loss functions have been proposed to approximate it \cite{bartlett2006convexity,masnadi2009design,nock2009efficient}, such as ramp loss \cite{brooks2011support} and unhinged loss \cite{van2015learning}, which are also proved to be robust to label noise under certain conditions \cite{ghosh2015making}. Specifically, \cite{manwani2013noise} formally defines a ``noise-tolerant" robust loss if the minimization under it with noisy labels would achieve the same solution as that with noise-free labels. \cite{ghosh2017robust} further relaxes the definition with loss bounded conditions to make the definition useful in guiding construction of rational robust losses in practice. Inspired by this formulation, several losses have been designed very recently and proved to satisfy loss bounded conditions, like the GCE \cite{zhang2018generalized} and SL \cite{wang2019symmetric}, both originated from the CE loss by introducing some noise-robust factor.

Although these robust loss functions help improve the robustness of a learning algorithm on noisy labels, they still have evident deficiencies in practice. On the one hand, they inevitably involve hyperparameter(s) to control their robustness extents against different noise rates. These hyperparameters need to be manually preset or tuned by heuristic strategies like cross-validation, naturally conducting efficiency issue and difficulty in practical implementations. On the other hand, the relatively complex form of robust loss will increase the non-convexity of the objective function used for training the network. Together with the non-convexity brought by the complicated network architectures, the minimization with all involved network parameters is highly non-convex, making the problem easily trapped to an unexpected solution with poor generalization capability even under properly preset hyperparameters in the robust loss. It is thus still challenging to construct a generally useful algorithm for the problem.

To alleviate the aforementioned issues, this paper presents an adaptive hyperparameter learning strategy to automatically tune the hyperparameter and thus learn the robust loss from data. Specifically, this study mainly made three-fold contributions.
 \begin{itemize}
\item The proposed algorithm realizes a mutual amelioration between automatically tuning proper hyperparameters involved in a robust loss and learning suitable network parameters. To the best of our knowledge, this is the first work to handle the robust loss (with explicit and concise forms) adaptive learning under noisy labels.
\item Four kinds of STOA robust loss functions, including GCE \cite{zhang2018generalized}, SL \cite{wang2019symmetric}, Bi-Tempered \cite{amid2019robust} and PolySoft \cite{gong2018decomposition}, are attempted to be integrated into our algorithm,
    showing the generality of our algorithm on adaptive robust loss learning with noisy labels. Especially, besides GCE and SL, which have been proved to be theoretically robust under loss bounded condition, we also prove that the loss bounded conditions for Bi-Tempered and PolySoft, implying their intrinsic robustness. These robust losses are thus all with sound theoretical guarantee and potentially useful.
\item Comprehensive experiments substantiate the effectiveness of the proposed algorithm, especially its superiority beyond conventional hyperparameter setting strategy. Specifically, from the experiments, it is interestingly seen that through iteratively ameliorating both robust loss hyperparameters and deep network parameters, our algorithm is capable of exploring good solution for the problem with evidently better generalization than that extracted by conventional hyperparameter tuning strategy with even carefully tuned hyperparameters. This might show a new potential way for exploring solutions with better generalization for such highly non-convex robust learning problems.

 \end{itemize}

The paper is organized as follows. Section \ref{related_work} reviews the related works. Section \ref{ARL_algo} introduces the four robust loss forms used in this paper, and proves the theoretical robustness theory for Bi-Tempered and PolySoft. Our main algorithm for adaptive robust loss learning is also presented in this section. Experiments are demonstrated in Section \ref{experiment}, and a conclusion is finally made.
\vspace{-1mm}
\section{Related Work} \label{related_work}\vspace{-1mm}
\textbf{Deep learning with noisy labels.} There are various approaches raised for handling robust learning issues under noisy labels, which can be roughly divided into four categories: label correction, loss correction, sample reweighting and robust loss setting.

The label correction approach aims to correct noisy labels to their true ones via a supplemental clean label inference step, characterized by directed graphical models \cite{xiao2015learning}, conditional random fields~\cite{vahdat2017toward} or knowledge graphs \cite{li2017learning}. Comparatively, the loss correction approach assumes that there exists a noise transition matrix defining the probability of one class changed to another. Typically, \cite{sukhbaatar2014training,goldberger2016training} modeled the matrix using a linear layer on the top of the DNNs, Forward \cite{patrini2017making} used the noise transition matrix to modify the loss function, and GLC \cite{hendrycks2018using} used additional meta data to estimate the transition matrix.

\textbf{Robust loss approach.}
Based on the noise-tolerant definition given by \cite{natarajan2013learning}, it has been proven that 0-1 loss, sigmoid loss, ramp loss, and probit loss are all noise-tolerant under some conditions \cite{ghosh2015making}. \cite{ghosh2015making} further relaxed this definition as bounded loss condition to make the theory better feasible in practice. Recently, curriculum loss, a tighter upper bound of the 0-1 loss was proposed in \cite{lyu2019curriculum}, which can adaptively select samples for training as a curriculum learning process. Generalized to multi-class problem, MAE (Mean Absolute Error) is proved to be robust to  symmetric label noise and class-conditional noise.

Under the advanced noise-robust understanding provided by \cite{ghosh2017robust}, some new robust losses have been designed on the basis of classical CE loss very recently expected to be finely performed in real practice. Zhang et al. \cite{zhang2018generalized}  demonstrated that it is hard to train DNN with MAE, and proposed to combine MAE and CE losses to obtain a new loss function, GCE, which behaves very like a weighted MAE, to handle the noisy label issue. \cite{wang2019symmetric} observed that the learning procedure of DNNs with CE loss is class-biased, and proposed a Reverse Cross Entropy (RCE) to help robust learning. Besides, \cite{amid2019robust} also presented a robust loss called Bi-Tempered by introducing two tunable temperatures to the traditional softmax layer and CE loss, which makes the loss be bounded and heavy-tail. Xu et al., \cite{xu2019l_dmi}  also provided a novel information-theoretic robust loss function different from the distance-based loss as aforementioned.

\textbf{Sample reweighting approach.} The main idea of this approach is to assign weights to losses of all training samples, and iteratively update these weights based on the loss values during the training process \cite{kumar2010self,jiang2014easy,zhao2015self}. Such a loss-weight function is generally set as monotonically decreasing, enforcing a learning effect that samples with larger loss, more possible to be noisy labeled as compared with small loss samples, are with smaller weights to suppress their effect to training. In this manner, the negative influence of noisy labels can be possibly alleviated. An interesting result is that when this monotonically decreasing weighting function makes this re-weighting learning process equivalent to solving an implicit robust loss function \cite{meng2017theoretical}, which constructs a close relationship between this strategy with robust loss approach. Very recently, some advanced sample reweighting methods have been raised inspired by the idea of meta-learning \cite{shu2019meta}, which possesses a much more complicated weighting scheme to the conventional reweighting strategies. This makes them able to deal with more general data bias cases other than noisy labels, like class imbalance. The fine theoretical basis, like noise tolerant, however, is also lost and almost impossible to be further prompted due to their complex implementation formats.

\textbf{Learning adaptive loss.} Some other methods have also attempted to directly learn a good proxy for an underlying evaluation loss. For example, learning to teach \cite{wu2018learning} dynamically learned the loss through a teacher network outputting the coefficients matrix of the general loss function. Xu et al., \cite{xu2018autoloss}  learned a discrete optimization schedule that alternates between different loss functions at different time-points. Adaptive Loss Alignment \cite{huang2019addressing} extended work in \cite{wu2018learning} to loss-metric mismatch problem. \cite{grabocka2019learning} tried to learn the surrogate losses for non-differentiable and non-decomposable loss function. These methods, however, generally attain losses with complicated forms, making them hardly to be theoretically analyzed with robust loss theory.

\textbf{Hyperparameter optimization.} Hyperparameter optimization was historically investigated by selecting proper values for each hyperparameter to obtain better performance on validation set. Typical methods include grid search, random search \cite{bergstra2012random}, Bayesian optimization \cite{snoek2012practical,swersky2013multi}, etc. Recently, meta-learning based strategy has been gradually more investigated \cite{franceschi2018bilevel,franceschi2017forward,maclaurin2015gradient,pedregosa2016hyperparameter}. This paper can be seen as a specific exploration of this methodology on adaptive robust loss learning issue on noisy labels.
\vspace{-1mm}
\section{Adaptive Robust Loss Learning} \label{ARL_algo}\vspace{-1mm}
\subsection{Preliminaries}\vspace{-1mm}
We consider the problem of $c$-class classification. Let $\mathcal{X} \subset \mathbb{R}^d$ be the feature space, and $\mathcal{Y}=\{1,2,\cdots,c\}$ be the label space. Assume the DNN architecture is with a softmax output layer.
Denote the network as a function with input $\mathbf{x}\in \mathcal{X}$ and output as $f(\mathbf{x};\mathbf{w})$, where $f: \mathcal{X}\rightarrow \mathbb{R}^c$, where
$\mathbf{w}$ represent the network parameters. $f_j(\mathbf{x};\mathbf{w})$, representing the $j$-th component ($j=1,\cdots,c$) of $f(\mathbf{x};\mathbf{w})$, then satisfies $\sum_{j=1}^{c} f_j(\mathbf{x};\mathbf{w}) =1, f_j(\mathbf{x};\mathbf{w})\geq 0$.  Given training data $D = \{(\mathbf{x}_i, \mathbf{y}_{i})\}_{i=1}^{N} \in(\mathcal{X} \times \mathcal{Y})^N$, for any loss function $\mathcal{L}$, the (empirical) risk of the network classifier is defined as:
 \begin{align*}
 \mathcal{L}(D,\mathbf{w}) = \frac{1}{N}\sum_{i=1}^N \mathcal{L}(f(\mathbf{x}_i;\mathbf{w}),\mathbf{y}_i).
 \end{align*}
The commonly used CE loss can be written as:
\begin{align}
\mathcal{L}_{CE}(D,\mathbf{w})= -\frac{1}{N}\sum_{i=1}^N \sum_{j=1}^c y_{ij} \log f_j(\mathbf{x}_i;\mathbf{w}), \label{LCE}
\end{align}
where $y_{ij}$ denotes the $j$-th component of $\mathbf{y}_i$. Generally in all components of $\mathbf{y}_i$, only one is $1$ and all others are $0$.

\subsection{Typical Robust Loss Forms}
We first introduce the forms of some typical robust losses.

\textbf{Generalized Cross Entropy (GCE).} To exploit the benefits of both the noise-tolerant property of MAE and the implicit weighting scheme of CE for better learning, Zhang et al., \cite{zhang2018generalized} proposed the GCE loss as follows:
\begin{align}\label{eqgce}
\begin{split}
\mathcal{L}_{GCE}(D,\mathbf{w};{\color{red} q})  &= \frac{1}{N} \sum_{i=1}^{N}\frac{(1- f_{j_i}(\mathbf{x}_i)^{\color{red} q})}{{\color{red} q}},
\end{split}
\end{align}
where $j_i$ denotes the $j$'s index of the term $y_{ij}=1$ for each $i$, and $q\in (0,1]$. GCE loss degenerates to the CCE when $q$ approaches to 0 and becomes to MAE loss when $q = 1$.

\textbf{Symmetric Cross Entropy (SL).} Wang et al. \cite{wang2019symmetric} proposed an extra term for CE to make it noise tolerant and designed the Reverse Cross Entropy (RCE):
\begin{align*}
\mathcal{L}_{RCE} = -\frac{1}{N}\sum_{i=1}^N \sum_{j\neq j_i} Af_j(\mathbf{x}_i;\mathbf{w}),
\end{align*}
where $A<0$ is a preset constant. The SL loss is defined as:
\begin{align}\label{eqsl}
\mathcal{L}_{SL}(D,\mathbf{w};{\color{red}\gamma_1,\gamma_2)} = {\color{red}\gamma_1}\mathcal{L}_{CE} +{\color{red}\gamma_2}\mathcal{L}_{RCE}.
\end{align}

\textbf{Bi-Tempered logistic Loss (Bi-Tempered).} Amid et al. \cite{amid2019robust} replaced the logarithm and exponential of the logistic loss with corresponding ``tempered'' versions function $\log_t,\exp_t$, to make the loss functions bounded to handle large-margin outliers and softmax function heavy-tailed to handle small-margin mislabeled examples. Specifically, they define $
\log_t(x)=\frac{1}{1-t}(x^{1-t}-1)
$
 and
$
\exp_t(x)=[1+(1-t)x]_+^{1/(1-t)},
$
where $[\cdot]_+=\max\{\cdot,0\}$. The Bi-Tempered Loss function is then defined as~\cite{amid2019robust}:
\begin{align}\label{eqbi}
\begin{split}
\mathcal{L}_{Bi}(D,\mathbf{w};{\color{red}t_1,t_2})& = -\frac{1}{N}\sum_{i=1}^N [\log_{{\color{red}t_1}}\hat{f}_{j_i,{\color{red}t_2}}(\mathbf{x}_i) \\
&+\frac{1}{2-{\color{red}t_1}}(1-\sum_{j=1}^c\hat{f}_{j,{\color{red}t_2}}(\mathbf{x}_i)^{2-{\color{red}t_1}})],
\end{split}
\end{align}
where $0\leq t_1<1, t_2>1$, $\hat{f}_{j,t} =\exp_{t} (z_j-\gamma_t(\mathbf{z}))$, $z_j$ is the input of softmax layer, %calculated by $f_j(\textbf{x}_i;\mathbf{w}) = \frac{e^{z_j}}{\sum_{k=1}^c e^{z_k}}$,
and $\gamma_t(\mathbf{z})$ is calculated by letting $\sum_{j=1}^c \exp_{t} (z_j-\gamma_{t}(\mathbf{z}))=1$.

\textbf{Polynomial Soft Weighting loss (PolySoft).} Self-paced learning (SPL) is a typical sample reweighting strategy to handle noisy labels by setting   monotonically decreasing weighting function \cite{kumar2010self,jiang2014easy,zhao2015self}. It has been proved that such  re-weighting learning process is equivalent to minimizing an latent robust loss \cite{meng2017theoretical}, and it thus can also be seen as a standard robust loss method. Recently, Gong et al. \cite{gong2018decomposition} proposed a polynomial soft weighting scheme for SPL, which can generally approximate monotonically decreasing weighting functions. By setting the CE loss as the basis loss form, the latent robust loss of this method is:
\begin{align}\label{eqlatent}
\begin{split}
&\mathcal{L}_{Poly}(D,\mathbf{w};{\color{red} \lambda,d}) = \\
&\begin{cases}
\frac{({\color{red}d}-1){\color{red}\lambda}}{{\color{red}d}}\left[1- (1-\frac{\mathcal{L}_{CE}(D,\mathbf{w})}{{\color{red}\lambda}})^{\frac{{\color{red}d}}{{\color{red}d}-1}}\right],& \mathcal{L}_{CE}<{\color{red}\lambda},\\
\frac{({\color{red}d}-1){\color{red}\lambda}}{{\color{red}d}},& \mathcal{L}_{CE} \geq {\color{red}\lambda},
\end{cases}
\end{split}
\end{align}
where $\mathcal{L}_{CE}$ is defined as in Eq. (\ref{LCE}).

\subsection{Adaptive Robust Loss Learning Algorithm}
It can be observed that all aforementioned robust loss functions contain hyperparameter(s), e.g., $q$ in $\mathcal{L}_{GCE}$ (Eq.(\ref{eqgce})), $\gamma_1,\gamma_2$ in $\mathcal{L}_{SL}$ (Eq.(\ref{eqsl})), $t_1,t_2$ in $\mathcal{L}_{Bi}$ (Eq.(\ref{eqbi})) and $\lambda,d$ in $\mathcal{L}_{Poly}$ (Eq.(\ref{eqlatent})). Instead of manually presetting or tuning them by cross-validation, we provide the following algorithm to adaptively learn these hyperparameter(s), by borrowing the idea of recent meta-learning techniques \cite{schmidhuber1992learning,thrun2012learning,finn2017model,franceschi2018bilevel,shu2018small,shu2019meta}.

\textbf{The Meta-learning Objective.} Given training dataset $D$, the net parameters are trained by optimizing the following minimization problem under certain robust loss $\mathcal{L}_{Train}$:
\begin{align}\label{eqroust}
\mathbf{w}^*(\Lambda) = \arg\min_{\mathbf{w}} \mathcal{L}_{Train}(D,\mathbf{w};\Lambda)
\end{align}
where $\Lambda$ denotes the hyperparameter set of $\mathcal{L}_{Train}$.

\begin{algorithm}[t]
	\vspace{0mm}
	\renewcommand{\algorithmicrequire}{\textbf{Input:}}
	\renewcommand{\algorithmicensure}{\textbf{Output:}}
	\caption{The Adaptive Robust Loss (ARL) Algorithm}
	\label{alg1}
	\begin{algorithmic}[1]  \small
		% \STATE {\bfseries Input:} Training data $\mathcal{D}$, meta set $\mathcal{D}^{(m)}$, batch size $n,n^{(m)}$, max iterations $T$.
		%   \STATE {\bfseries Output:} Student model parameters $w^{(T)}$
		\REQUIRE  Training data $S$, meta data $S_{meta}$, batch size $n,m$, max iterations $T$.
		\ENSURE  Classifier network parameter $\mathbf{w}$, robust loss hyperparameter  $\Lambda$.
		%\REPEAT
		\STATE Initialize classifier network parameter $\mathbf{w}^{(0)}$ and robust loss $\mathcal{L}_R$ hyperparameter  $\Lambda^{(0)}$.
		\FOR{$t=0$ {\bfseries to} $T-1$}
		\STATE $\{x,y\} \leftarrow$ SampleMiniBatch($S,n$).
		\STATE $\{x^{(m)},y^{(m)}\} \leftarrow$ SampleMiniBatch($S_{meta},m$).
		\STATE Update $\Lambda^{(t)}$ by Eq. (\ref{eqlambda}).
		\STATE Update $\mathbf{w}^{(t)}$ by Eq. (\ref{eqpara}).
		\ENDFOR
		%\UNTIL{$noChange$ is $true$}
	\end{algorithmic}
\end{algorithm}

Our method aims to automatically learn the hyperparameters $\Lambda$  in a meta-learning manner \cite{finn2017model,ren2018learning,shu2019meta}. Specifically, assume that we have a small amount meta-data set (i.e., with clean labels) $D_{meta} = \{x_i^{(m)},y_i^{(m)}\}_{i=1}^M$, representing the meta-knowledge of ground-truth sample-label distribution, where $M$ is the number of meta-samples, and $M\ll N$. We can then formulate a meta-loss minimization problem with respect to $\Lambda$ as:
\begin{align} \label{eqmeta}
\Lambda^* = \arg\min_{\Lambda} \mathcal{L}_{Meta}(D_{meta},\mathbf{w}^*(\Lambda)),
\end{align}
where $\mathcal{L}_{Meta}$ represents the loss imposed on meta data. Since meta data are all clean, it is employed as the conventional loss forms without hyperparameter, like CE loss.

\textbf{Learning Algorithm.} Calculating the optimal $\mathbf{w}^*$ and $\Lambda^*$ requires two nested loops of optimization, which is expensive to obtain the exact solution \cite{franceschi2018bilevel}. Here we adopt an online approximation strategy \cite{finn2017model, shu2019meta} to jointly update both sets of parameters in an iterative manner to guarantee the efficiency of the algorithm.

At iteration step $t$, we need to update hyperparameter $\Lambda^{(t)}$ on the basis of the net parameter $\mathbf{w}^{(t-1)}$ and hyperparameter $\Lambda^{(t-1)}$ obtained in the last iteration by minimizing the meta loss defined in Eq.(\ref{eqmeta}). To guarantee efficency and general feasibility, SGD is employed to optimize the parameters on $m$ mini-batch samples $D_m$ from $D_{meta}$, i.e.,
\begin{align} \label{eqlambda}
\Lambda^{(t)} = \Lambda^{(t-1)} -  \beta\nabla_{\Lambda} \mathcal{L}_{Meta} (D_m, \tilde{\mathbf{w}}^{(t)}(\Lambda))\Big|_{\Lambda^{(t-1)}},
\end{align}
where the following equation is used to formulate $\tilde{\mathbf{w}}^{(t)}((\Lambda))$ on a mini-batch training samples $D_n$ from $D$:
\begin{align}
\tilde{\mathbf{w}}^{(t)}(\Lambda) = \mathbf{w}^{(t-1)} - \alpha  \nabla_{\mathbf{w}} \mathcal{L}_{Train}(D_n,\mathbf{w};\Lambda)\Big|_{\mathbf{w}^{(t-1)}},
\end{align}
which is inspired from MAML \cite{finn2017model}, and $\alpha,\beta$ is the step size.

When obtained the parameter $\Lambda^{(t)}$, the network parameters $\mathbf{w}^{(t)}$ can then be updated by:
 \begin{align}\label{eqpara}
 \mathbf{w}^{(t)}= \mathbf{w}^{(t-1)} - \alpha \nabla_{\mathbf{w}}\mathcal{L}_{Train}(D_n,\mathbf{w};\Lambda^{(t)})\Big|_{\mathbf{w}^{(t-1)}}.
 \end{align}

The Adaptive Robust Loss (ARL) Algorithm can then be summarized in Algorithm \ref{alg1}. All computations of gradients can be efficiently implemented by automatic differentiation techniques and easily generalized to any deep learning architectures. The algorithm can be easily implemented using popular deep learning frameworks like PyTorch \cite{paszke2017automatic}. The algorithm can then be easily integrated with any robust loss to make their hyperparameter automatically learnable. Specifically, we denote the ARL algorithms on robust losses defined in Eq.(\ref{eqgce}),(\ref{eqsl}),(\ref{eqbi}) and (\ref{eqlatent}) as A-GCE, A-SL, A-Bi-Tempered and A-PolySoft, respectively.

\begin{table*}[t]
	\caption{Test accuracy (\%) of all competing methods on CIFAR-10 and CIFAR-100 under different noise rates. The best results are in bold.}\label{table1} \vspace{1mm}
	\centering
	\begin{footnotesize}
		\begin{tabular}{c|c|c|c|c|c|c|c|c}
			\toprule
			\multirow{3}{*}{Models}             & \multirow{3}{*}{Datasets}              &  \multirow{3}{*}{Methods}    &\multicolumn{4}{c|}{Symmetric Noise}   & \multicolumn{2}{c}{Asymmetric Noise} \\
			\cline{4-9}
			&                      &            &  \multicolumn{4}{c|}{Noise Rate $\eta$}         & \multicolumn{2}{c}{Noise Rate $\eta$}   \\
			\cline{4-9}
			&                     &            &        0            &        0.2    &       0.4         &     0.6             &     0.2      &       0.4                   \\	
			\hline
			\hline
			\multirow{24}{*}{ResNet-32} &\multirow{12}{*}{CIFAR-10} & CE  &     92.89$\pm$0.32  & 76.83$\pm$2.30 & 70.77$\pm$2.31   &  63.21$\pm$4.22     & 76.83$\pm$2.30  &  70.77$\pm$2.31                \\
			&                       &Forward  &    \textbf{93.03$\pm$0.11}   & 86.49$\pm$0.15 & 80.51$\pm$0.28   &  75.55$\pm$2.25     & 87.38$\pm$0.48 &     78.98$\pm$0.35                             \\
			&                            &DMI &   90.91$\pm$0.20    & 87.59$\pm$0.21 & 85.13$\pm$0.10   &   80.23$\pm$0.39    &    89.08$\pm$0.49  &  79.33$\pm$0.65                          \\
			&             &  Meta-Weight-Net  &      92.04$\pm$0.15 &89.19$\pm$0.57  & 86.10$\pm$0.18   & 81.31$\pm$0.37      &      90.33$\pm$0.61  &  87.54$\pm$0.23         \\
			&                    &PolySoft    &      91.40$\pm$0.39 &  87.53$\pm$0.48 & 81.49$\pm$0.34  &  75.87$\pm$0.25     &  85.99$\pm$1.77 &82.71$\pm$0.99       \\
			&    & {\color{blue}A-PolySoft}   &      92.12$\pm$0.12 & \textbf{89.73$\pm$0.20} & \textbf{87.22$\pm$0.36}   & \textbf{82.49$\pm$0.30}      &  \textbf{90.41$\pm$0.16} &  \textbf{87.75$\pm$0.23 }                    \\
			&                            & GCE&      90.03$\pm$0.30 &88.51$\pm$0.37  &85.48$\pm$0.16    &  81.29$\pm$0.23     &   88.55$\pm$0.22   &     83.31$\pm$0.14        \\
			&            & {\color{blue}A-GCE}&    91.47$\pm$0.19   &89.07$\pm$0.27  & 86.36$\pm$0.14   &   81.64$\pm$0.11     &   89.51$\pm$0.07   &86.35$\pm$0.17 \\
			&                            & SL &    89.37$\pm$0.13   & 88.76$\pm$ 0.56 & 85.84$\pm$0.74  &   81.38$\pm$1.39     &   87.63$\pm$0.34  &   83.48$\pm$0.48              \\
			&            & {\color{blue}A-SL} &   91.50$\pm$0.16    &  89.53$\pm$0.22 &  86.36$\pm$0.41 &   82.19$\pm$0.30     &   89.54$\pm$0.28   &  86.45$\pm$0.20         \\
			&                    & Bi-Tempered&   90.11$\pm$0.23    & 88.51$\pm$0.31  & 84.93$\pm$0.67  & 77.82$\pm$0.79       &   88.23$\pm$ 0.23  &    82.43$\pm$0.23       \\
			&  & {\color{blue}A-Bi-Tempered}    &    92.24$\pm$0.20   & 89.37$\pm$0.09  &  86.32$\pm$0.28 & 81.70$\pm$0.21       &   89.88$\pm$0.30   &  86.86$\pm$0.28        \\ \cline{2-9}\cline{2-9}
			&\multirow{12}{*}{CIFAR-100} &  CE&  \textbf{70.50$\pm$0.12 }    &  50.86$\pm$0.27&43.01$\pm$1.16    & 34.43$\pm$0.94       &   50.86$\pm$0.27  &  43.01$\pm$1.16           \\
			&                        & Forward&  67.81$\pm$0.61     &63.75$\pm$0.38  & 57.53$\pm$0.15   & 46.44$\pm$1.03       &    64.28$\pm$0.23 & 57.90$\pm$0.57               \\
			&                            & DMI&   68.40$\pm$0.23    &  62.66$\pm$0.05& 56.95$\pm$0.11   &  46.30$\pm$0.10     &   64.05$\pm$0.18   &   58.08$\pm$0.22            \\
			&             &Meta-Weight-Net    &     69.13$\pm$0.33     & 64.22$\pm$0.28 &   58.64$\pm$0.47&    47.43$\pm$0.76     &  64.22$\pm$0.28 &58.64$\pm$0.47           \\
			&                    &  PolySoft  &    68.26$\pm$0.25     &  62.41$\pm$0.38   &  56.16$\pm$0.30 &    45.23$\pm$0.47   &   63.05$\pm$0.61  &  56.09$\pm$ 0.26             \\
			&   &{\color{blue}A-PolySoft}     &    68.92$\pm$0.41   & \textbf{65.37$\pm$1.43} & \textbf{ 61.38$\pm $0.47} & \textbf{ 52.23$\pm$0.63 }     &  \textbf{64.42$\pm$0.26}  & \textbf{58.73$\pm$0.17 }             \\
			&                            & GCE &   67.39$\pm$0.12     &  63.97$\pm$0.43 &58.33$\pm$0.35   &  41.73$\pm$0.36      & 62.07$\pm$0.41 & 55.25$\pm$0.09               \\
			&              &{\color{blue}A-GCE}&    67.57$\pm$0.32     &  64.58$\pm$0.30  & 58.50$\pm$0.15  &  42.16$\pm$0.63   &  62.46$\pm$ 0.52 &56.75$\pm$0.44          \\
			&                            & SL   &  66.43$\pm$0.43    & 52.46$\pm$0.18 &  51.28$\pm$0.73  & 38.39$\pm$1.53      &  52.04$\pm$0.89& 44.01$\pm$1.91               \\
			&             & {\color{blue}A-SL} & 68.07$\pm$0.51    & 63.73$\pm$0.27 &  57.99$\pm$0.37  &   45.75$\pm$0.66     & 63.25$\pm$0.33 &   56.83$\pm$0.19              \\
			&                     & Bi-Tempered&   67.68$\pm$0.25   &  63.45$\pm$0.48&  57.25$\pm$0.16 &    44.72$\pm$0.39     &   63.12$\pm$0.28 & 55.37$\pm$0.56               \\
			&      &{\color{blue}A-Bi-Tempered}& 69.32$\pm$0.19    &   64.48$\pm$0.53& 59.26$\pm$0.12 &    48.62$\pm$0.32      &   63.78$\pm$0.27  & 56.56$\pm$0.08           \\
			\bottomrule
		\end{tabular} \vspace{-2mm}
	\end{footnotesize}
\end{table*}

\subsection{Noise-robust Properties of Utilized Losses}
It can be seen from Fig.\ref{fig1} the adopted GCE, SL, Bi-Tempered and PolySoft losses are all robust amelioration from CE loss to noisy labels. All of them tend to be flat when loss becomes larger so as to suppress the negative influence by large losses brought by noisy samples with incorrect labels. In theory, actually they all satisfy loss bounded conditions, and are noise-tolerant under certain conditions.
This property has been proved for GCE \cite{zhang2018generalized} and SL \cite{wang2019symmetric}, and we then provide the related results for the other two.

Denote the true label of $x$ as $\hat{y}$, in contrast to its noisy label $y$, and $D_c$ and $D_n$ as the underlying distributions of clean and noisy data, respectively. Let $R_{\mathcal{L}}(f)=\mathbb{E}_{D_c} [\mathcal{L}(f(x),\hat{y})]$ be the risk of classifier $f$ under clean labels, and  $R_{\mathcal{L}}^{\eta}(f)=\mathbb{E}_{D_n} [\mathcal{L}(f(x),y)]$ as the risk of classifier $f$ under label noise rate $\eta$.
A loss function $\mathcal{L}$ is defined to be noise tolerant \cite{manwani2013noise,ghosh2017robust} if $\hat{f}$ on noisy data has the same misclassification probability
as that of $f^*$ on clean data, where $\hat{f}$ and $f^*$ are the global minimizers of $R_{\mathcal{L}}^{\eta}(f)$ and $R_{\mathcal{L}}(f)$, respectively.
To make this definition more feasible in practice, Ghosh et al. \cite{ghosh2017robust} further relaxed this definition as the bounded loss condition and proved that under it the loss also possesses certain robustness capacity in theory  \cite{zhang2018generalized}, as proved for GCE \cite{zhang2018generalized} and SL \cite{wang2019symmetric}. We can further prove that the PolySoft and Bi-Tempered also possess these properties, as provided in the following theorems. The detailed proofs are listed in the supplementary material.
\begin{theorem}\label{TH1}
	Under the symmetric noise with $\eta \leq 1-\frac{1}{c}$, and $\lambda \geq \log c, d>1$, the PolySoft loss in Eq. (\ref{eqlatent}) satisfies
	\begin{align*}
	0\leq R_{\mathcal{L}}^{\eta}(f^*)-R_{\mathcal{L}}^{\eta}(\hat{f}) \leq A,
	A' \leq R_{\mathcal{L}}(f^*)-R_{\mathcal{L}}(\hat{f})\leq 0,
	\end{align*}
	where $A = \frac{c(d-1)\eta}{d(c-1)}(\lambda-\log c)\geq 0$, $A' = \frac{c(d-1)\eta}{d(c-1-\eta c)}[\log c -\lambda]<0$. Especially, when $\lambda = \log c$, we have $R_{\mathcal{L}}^{\eta}(f^*)=R_{\mathcal{L}}^{\eta}(\hat{f})$.
\end{theorem}
The theorem clarifies that the PolySoft loss is with loss bounded condition, and noise tolerant when $\lambda = \log c$.

\begin{theorem}\label{TH2}
	Under the symmetric noise with $\eta \leq 1-\frac{1}{c}$, and $ 0\leq t_1<1, t_2>1$, the Bi-Tempered loss in Eq.(\ref{eqbi}) satisfies
	\begin{align*}
	0\leq R_{\mathcal{L}}^{\eta}(f^*)-R_{\mathcal{L}}^{\eta}(\hat{f}) \leq A,
	A' \leq R_{\mathcal{L}}(f^*)-R_{\mathcal{L}}(\hat{f})\leq 0,
	\end{align*}
	where $A = \frac{\eta}{1-t_1}-\frac{\eta(c-c^{t_1})}{(c-1)(1-t_1)(2-t_1)}>0$, $A' =\frac{\eta(c-c^{t_1})}{(c-1-\eta c)(1-t_1)(2-t_1)} -\frac{\eta(c-1)}{(1-t_1)(c-1-\eta c)}<0$.
\end{theorem}
This theorem illustrates that the Bi-Tempered loss satisfies loss bounded condition. Albeit not noise tolerant, it still possesses certain theoretical robustness \cite{zhang2018generalized}.

\vspace{-1mm}
\section{Experimental Results} \label{experiment}\vspace{-1mm}
To evaluate the capability of the ARL algorithm, we implement experiments on CIFAR-10, CIFAR-100, TinyImageNet, as well as a
large-scale real-world noisy dataset Clothing1M.
\vspace{-1mm}
\subsection{Experimental Setup}\vspace{-1mm}
\textbf{Datasets.} We first verify the effectiveness of our method on two benchmark datasets: CIFAR-10 and CIFAR-100 \cite{krizhevsky2009learning}, consisting of $32\times32$ color images arranged in 10 and 100 classes, respectively. Both datasets contain 50,000 training and 10,000 test images. We random select 1,000 clean images in the validation set as meta data. Then, we verify our method on a larger and harder dataset called Tiny-ImageNet (T-ImageNet briefly), containing 200 classes with 100K training, 10K validation, 10K test images of $64\times64$. We random sample 10 clean images per class as meta data. These datasets are popularly used for evaluation of learning with noisy labels in the previous literatures \cite{reed2014training,patrini2017making,goldberger2016training}.

\textbf{Noise setting.} We test two types of label noise: symmetric noise and asymmetric (class-dependent) noise. \textbf{Symmetric} noisy labels are generated by flipping the labels of a given proportion of training samples to one of the other class labels uniformly \cite{zhang2016understanding}. For \textbf{asymmetric} noisy labels, we use the setting in \cite{shu2019meta}, where the label of each sample is independently flipped to two classes with same probability. Also, we consider a more realistic \textbf{hierarchical} corruption in CIFAR-100 as described in \cite{hendrycks2018using}, which applies uniform corruption only to semantically similar classes.

\textbf{Baselines.} We compare ARL algorithm with the following state-of-art methods, and implement all methods with default settings in the original paper by PyTorch. 1) \textbf{CE}, which uses CE loss to train the DNNs on noisy datasets. 2) \textbf{Forward} \cite{patrini2017making},  which corrects the prediction by the label transition matrix. 3) \textbf{DMI }\cite{xu2019l_dmi}, which uses mutual information based robust loss to train the DNNs.
4) \textbf{Meta-Weight-Net }\cite{shu2019meta}, which uses a MLP net to learn the weighting function in a data-driven fashion, representing the SOTA sample weighting methods. 5) \textbf{PolySoft} \cite{gong2018decomposition}, 6) \textbf{GCE} \cite{zhang2018generalized}, 7) \textbf{SL} \cite{wang2019symmetric}, 8) \textbf{Bi-Tempered} \cite{amid2019robust} represent the STOA robust loss methods.
The meta-data in these methods are used as validation set for cross-validation to search the best hyperparameters except for Meta-Weight-Net.

%\begin{figure}
%	\centering
%	\includegraphics[width=0.9\linewidth]{fig/comp}
%	\caption{Comparion of test accuracies between ARL algorithm and re-trained model using the hyper-parameters learned at the last step by ARL algorithm (denoted by -Opt).}
%	\label{fig:comp}
%\end{figure}
\textbf{Network structure.} We use ResNet-32 \cite{he2016deep} as our classifier network models for CIFAR-10 and CIFAR-100 dataset, and a 18-layer Preact ResNet \cite{he2016deep} for T-ImageNet.

\textbf{Experimental setup.} We train the models with SGD, at an initial learning rate 0.1 and a momentum 0.9, a weight decay $1\times10^{-3}$ with mini-batch size 100.
%For all baseline methods, ResNet-32 models, the learning rate decays 0.1 at 40 epochs and 80 epochs for a total of 120 epochs; WRN-28-10 models, the learning rate decays 0.1 at 80 epochs and 100 epochs for a total of 120 epochs; Preact ResNet-18 models, the learning rate decays 0.1 at 30 epochs and 60 epochs for a total of 90 epochs. And the hyper-parameter
For our proposed methods, ResNet-32 models, the learning rate decays 0.1 at 40 epochs and 50 epochs for a total of 60 epochs; Preact ResNet-18 models, the learning rate decays 0.1 at 30 epochs and 60 epochs for a total of 90 epochs. We use SGD to optimize hyperparameters, and the learning rate setting is the same as the classifier for different experiments.

\textbf{Hyperparameter setting.} For the methods in PolySoft, GCE, SL, Bi-Tempered, we used the optimal hyperparameter in the original paper or carefully searched by cross-validation. For our method, those hyperparameters are automatically learned.
% We set the initial hyperparameters in our methods as: 1) A-PolySoft: $\eta^{(0)} = 5$ for CIFAR10, $\eta^{(0)} = 7$ for CIFAR100, $\eta^{(0)} = 10$ for T-ImageNet\footnote{The theoretical noise-robust value is $\eta = \log c$, while in practice, we need to set a higher value to avoid no samples involving training.}, and $t^{(0)}=2$; 2) A-GCE: $q^{(0)}=0.7$ ; 3) A-SL:$\alpha^{(0)}=\beta^{(0)}=1$; 4) A-Bi-Tempered: $t_1^{(0)}=0.7, t_2^{(0)}=1.5$.
\vspace{-1mm}
\subsection{Robustness Performance Evaluation}\vspace{-1mm}
\textbf{Results on  CIFAR-10 and CIFAR-100.}  The classification accuracies of CIFAR-10 and CIFAR-100 under symmetric and asymmetric noise are reported in Table \ref{table1} with 5 random runs.  As can be seen, our {\color{blue}ARL algorithm}  (in color blue) improves on the original algorithm  via a large margin for almost all noise rates and all datasets. Table \ref{table2} shows classification accuracies of more realistic hierarchical label corruption on CIFAR-100 dataset. Our ARL algorithm can also improve the accuracy of the original algorithm and A-PolySoft outperforms all other baselines methods.

\begin{table*}[t]
	\caption{Test accuracy (\%) of ResNet-32 on CIFAR-100 with hierarchical noisy labels. The best results are in bold.}\label{table2} \vspace{1mm}
	\centering
	\begin{footnotesize}
		\begin{tabular}{c|c|c|c|c|c|c|c}
			\toprule
			\multicolumn{2}{c|}{Methods} & CE             & Forward        & DMI              & Meta-Weight-Net             & PolySoft          & {\color{blue}A-PolySoft}       \\ \hline
			\multirow{2}{*}{Noise Rate $\eta$} & 0.2                  & 51.31$\pm$0.27 & 64.35$\pm$0.33 & 64.51$\pm$0.08    &  64.38$\pm$0.38  & 63.51$\pm$0.38   &  \textbf{65.42$\pm$0.15}         \\
			& 0.4             &45.23$\pm$1.16  & 59.74$\pm$0.19 &60.09$\pm$0.10     &    59.41$\pm$0.79              &    58.63$\pm$0.12     &       \textbf{60.46$\pm$0.18}             \\ \hline \hline
			\multicolumn{2}{c|}{Methods}& GCE              & {\color{blue}A-GCE}         & SL               & {\color{blue}A-SL}               & Bi-Tempered       & {\color{blue}A-Bi-Tempered} \\ \hline
			\multirow{2}{*}{Noise Rate $\eta$} & 0.2             &   62.72$\pm$0.36  &   63.31$\pm$0.40          &  56.38$\pm$0.50  &   63.30$\pm$0.19           & 63.45$\pm$0.20    &64.99$\pm$0.25  \\
			& 0.4        &58.03$\pm$0.81    & 58.46$\pm$0.33              &48.34$\pm$0.33      &57.94$\pm$0.71     & 57.90$\pm$0.18    &59.90$\pm$0.53\\
			\bottomrule
		\end{tabular}\vspace{-3mm}
	\end{footnotesize}
\end{table*}

\begin{table}
	\caption{Test accuracy (\%) on T-ImageNet under different noise fractions. The best results are in bold. }\label{tableT} \vspace{1mm}
	\centering
	\begin{footnotesize}
		\begin{tabular}{c|c|c|c|c|c|c}
			\toprule
			\multirow{3}{*}{Methods}    &\multicolumn{4}{c|}{Symmetric Noise}   & \multicolumn{2}{c}{Asymmetric Noise} \\
			\cline{2-7}
			&  \multicolumn{4}{c|}{Noise Rate $\eta$}         & \multicolumn{2}{c}{Noise Rate $\eta$}   \\
			\cline{2-7}
			&        0            &        0.2    &       0.4         &     0.6             &     0.2      &       0.4                   \\	
			\hline
			\hline
			CE  &     55.01 & 43.94   & 35.14     &  20.45     & 42.12    &  33.58                \\
			Forward  &  \textbf{ 55.29}   & 46.57   & 38.01     &  24.43     & 44.98    &  36.99                   \\
			DMI &   54.50   & 46.10   & 40.35     &   25.23    &   44.82  & 36.68                  \\
			MW-Net &   53.58        &48.31   & 43.33      & 28.23      & 45.17       &   37.72          \\
			PolySoft    &  52.18          &  46.86   &40.76  &    21.48         &    43.99    &       36.11                \\
			{\color{blue}A-PolySoft}   &  54.18         & \textbf{49.24}  &   \textbf{43.67 }        &\textbf{28.46  }           &  \textbf{48.65}     &  \textbf{ 40.50 }                  \\
			GCE&   53.11        &47.72   & 38.96    &   23.93            &  45.62      &  35.32         \\
			{\color{blue}	A-GCE}&  53.46          &  48.22      &  41.40       &   24.11        &   46.18       &   36.47         \\
			SL &    52.48     &    44.33    &     35.18    &    21.82      &   44.18         &             34.69               \\
			{\color{blue}	A-SL} &  53.34    &   48.99       &  38.29         &   22.39         &47.68          &   37.78          \\
			Bi-Tem&  52.09        &  45.90       &  35.36        &       21.32       &  44.14       & 34.37        \\
			{\color{blue}A-Bi-Tem}    &  54.22        & 46.67   & 37.36         &  22.10             & 46.91     &  35.43      \\
			\bottomrule
		\end{tabular}\vspace{-1mm}
	\end{footnotesize}
\end{table}

Combining Table \ref{table1} and \ref{table2}, it can be observed that: 1) PolySoft and Bi-Tempered's performance drop quickly as the noise rate exceeds 0.4, and A-PolySoft and A-Bi-Tempered improve the accuracy around 5\% on CIFAR-100 thereafter. 2) SL drops more quickly on CIFAR-100 than CIFAR-10, and A-SL improves the accuracy more evidently, over 10\% on CIFAR-100 with 20\% noise rate. 3) A-PolySoft outperforms the STOA sample reweighting method Meta-Weight-Net, possibly attributed to its monotonically decreasing form of robust loss, making it noise robust, as illustrated in Section \ref{understand}.

\textbf{Results on  T-ImageNet.} To verify our approach on a more complex scenario, we summarize in Table \ref{tableT} the test accuracy on T-ImageNet with different noise setting. As we can see, for both noise settings with different noise rates, A-PolySoft outperforms other baselines. Meanwhile, our ARL algorithm improves the original algorithm stably.

%One problem is that the final learned hyper-parameters whether or not are the optimal to improve performance. We use the hyper-parameters learned at the last step to re-train the model, and the results on CIFAR-10 of ResNet-32 are presented in Fig.\ref{fig:comp}. We find that the re-trained model drops more when increasing noise amounts, which implies the dynamic learning process improves learning performance.

\begin{table*}[t]
	\caption{Test accuracy (\%) of different models on real-world noisy dataset Clothing1M. The best results are in bold.}\label{Table4} \vspace{1mm}
	\centering
	\begin{footnotesize}
		\begin{tabular}{c|c|c|c|c|c|c|c|c|c|c|c|c}
			\toprule
			Methods & CE    & Forward  &  DMI  &  MN-Net  & PolySoft  & {\color{blue}A-PolySoft} &  GCE   & {\color{blue}A-GCE} &SL  & {\color{blue}A-SL}& Bi-Tem  &{\color{blue}A-Bi-Tem}   \\   \hline
			Accuracy & 68.94 &   70.83 &  72.46 &    73.72 &    69.96  &  73.76                 &   69.75  &     70.55           &  71.02    &  71.83       &  69.89    &      70.14             \\
			\bottomrule
		\end{tabular} \vspace{-2mm}
	\end{footnotesize}
\end{table*}

	\begin{figure}[t]
	\centering
	%\vspace{-1.3cm}
	\subfigcapskip=-1mm
	\subfigure[Sample weight distribution of Meta-Weight-Net]{
		\label{fig1a} %% label for first subfigure
		\includegraphics[width=0.48\textwidth]{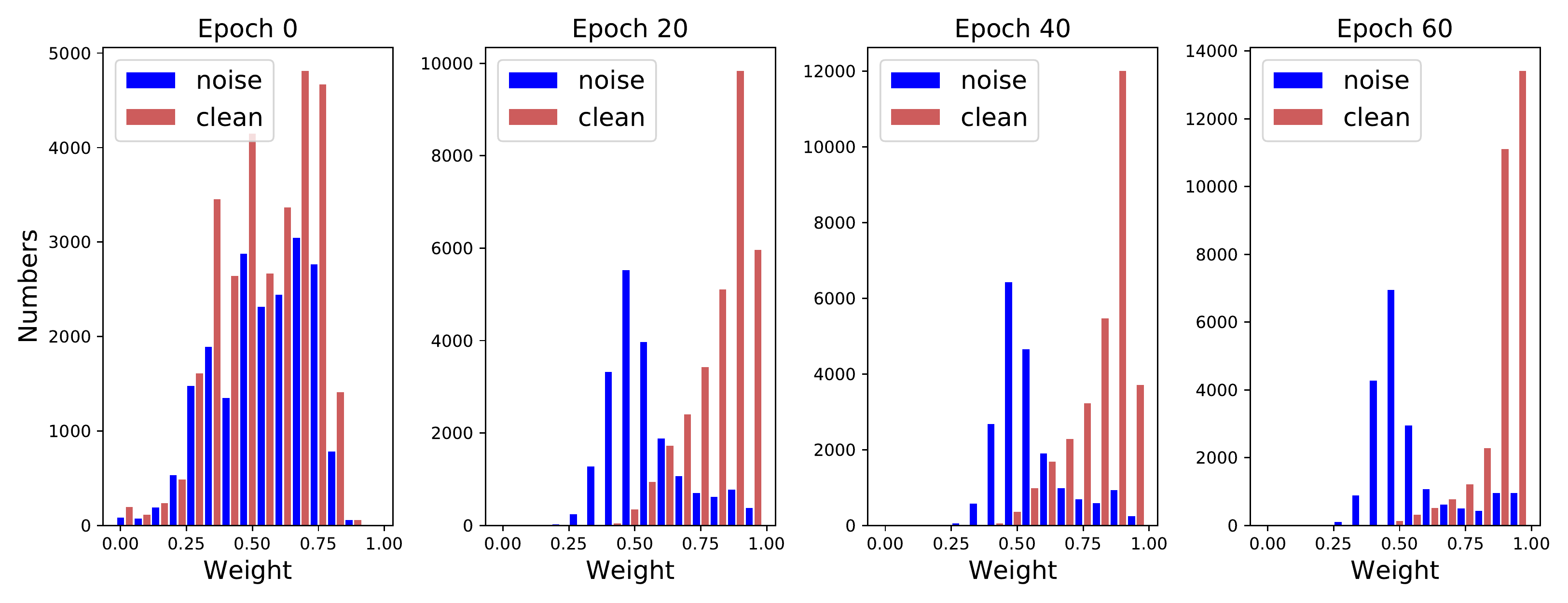}} \\ \vspace{-3mm}
	\subfigure[Sample weight distribution of A-PolySoft]{
		\label{fig1b} %% label for first subfigure
		\includegraphics[width=0.48\textwidth]{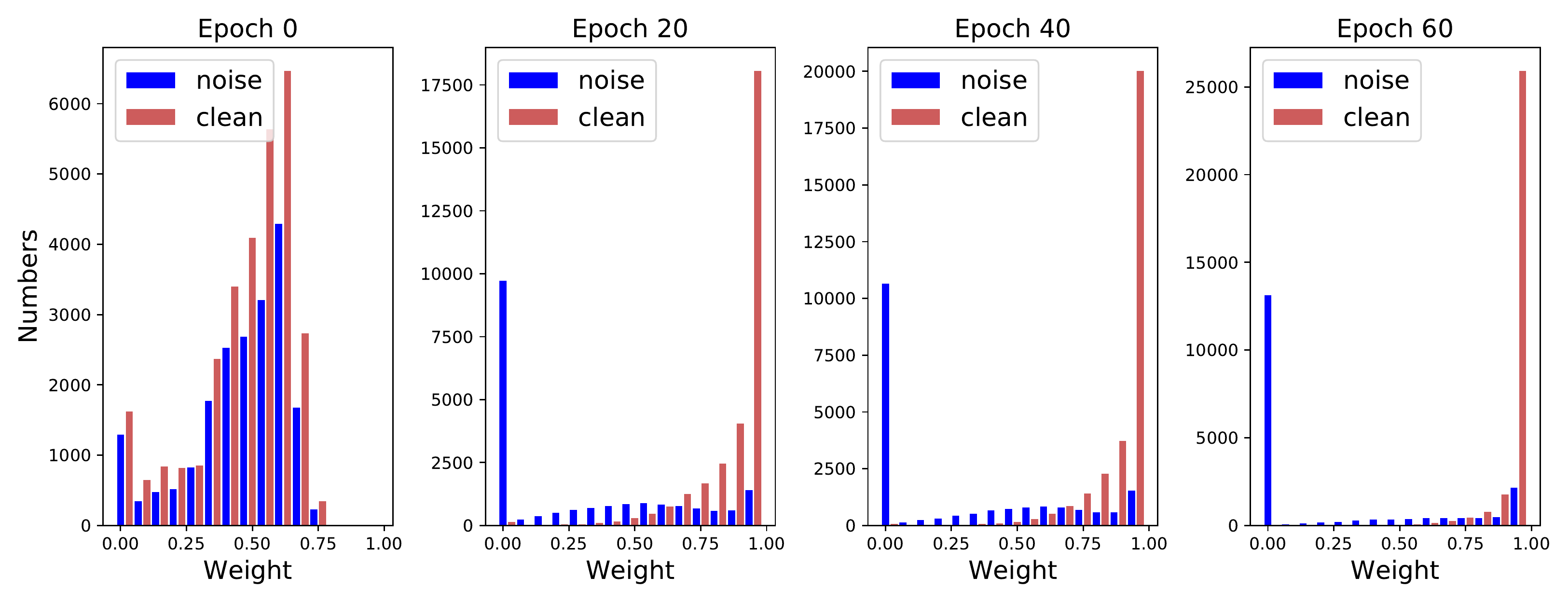}}
	\caption{Sample weight distribution on CIFAR-10 dataset under 40\% symmetric noise experiments during training process.  (a)-(b) present the sample weights produced by Meta-Weight-Net and A-PolySoft, respectively.}\label{fig2}
		\vspace{-4mm}
	\end{figure}

\vspace{-1mm}
\subsection{Towards Understanding of ARL Algorithm} \label{understand}\vspace{-1mm}
\textbf{How ARL adapt to noise extents.}
To understand how the ARL algorithm automatically fit noise extents, we plot the learned loss under different noise rates on CIFAR10 dataset in Fig.\ref{fig1}. It is easy to see that when the loss value is small, learned loss is almost the same as CE loss; while when loss becomes larger and exceeds a certain threshold, learned loss tends to be flat thereafter, behaving as suppressing the effect of samples with incorrect labels often with large losses. Furthermore, the learned loss tends to be flat earlier when the noisy rate is higher, which implies ARL algorithm can adjust loss function according to noise amounts to better encode and alleviate the noisy label effectly. Such noisy-rate adapting capability of our algorithm finely explains its superior robustness depicted in our experiments.

	\begin{figure}[t]
	\centering
	%\vspace{-1.3cm}
	\subfigcapskip=-1mm
	\subfigure[CE on clean data. ]{
		\label{fig1a} %% label for first subfigure
		\includegraphics[width=0.23\textwidth]{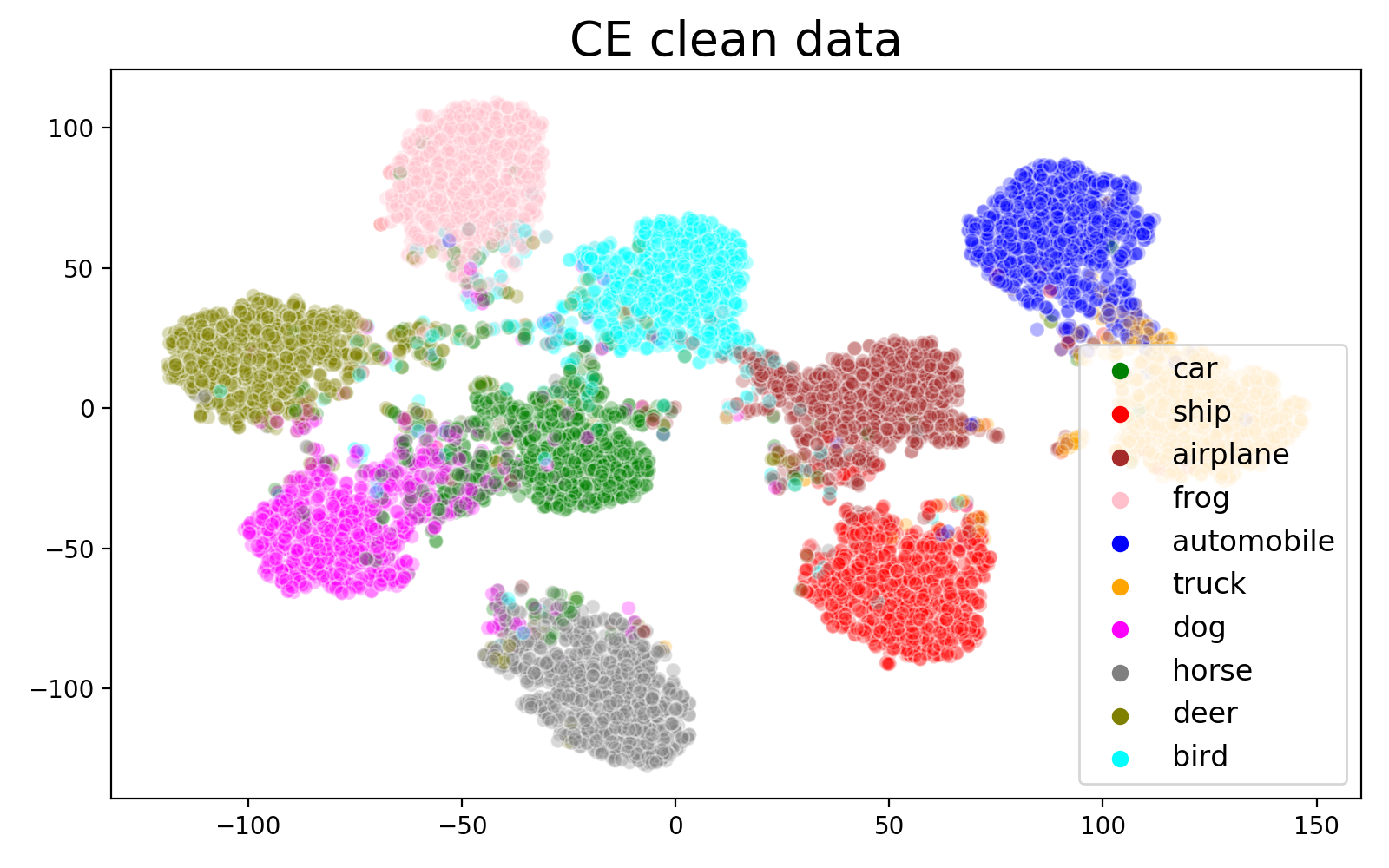}}
	\subfigure[CE on 60\% noise data.]{
		\label{fig1b} %% label for first subfigure
		\includegraphics[width=0.23\textwidth]{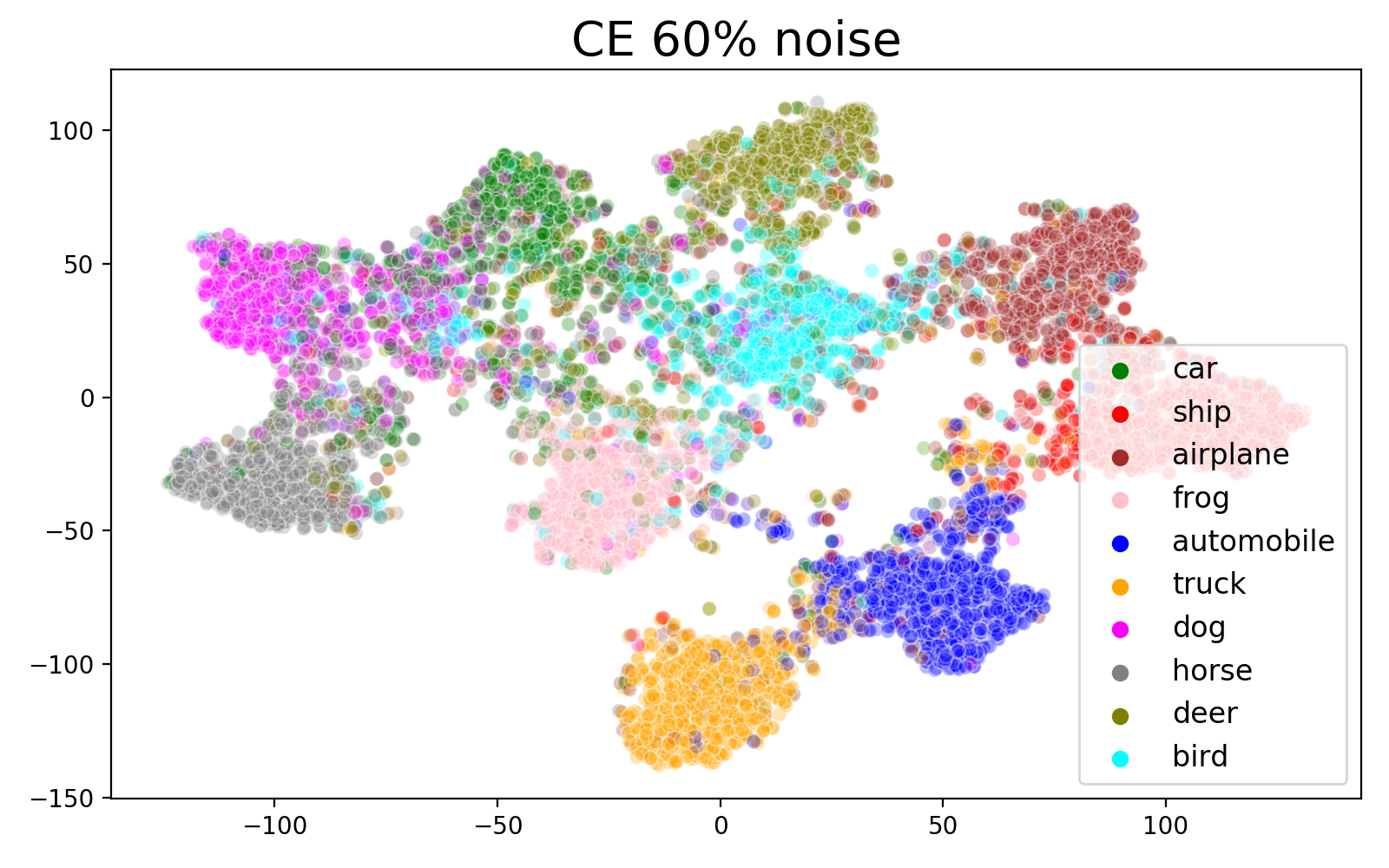}} \\ \vspace{-3mm}
	\subfigure[Bi-Tem on 60\% noise data. ]{
		\label{fig1a} %% label for first subfigure
		\includegraphics[width=0.23\textwidth]{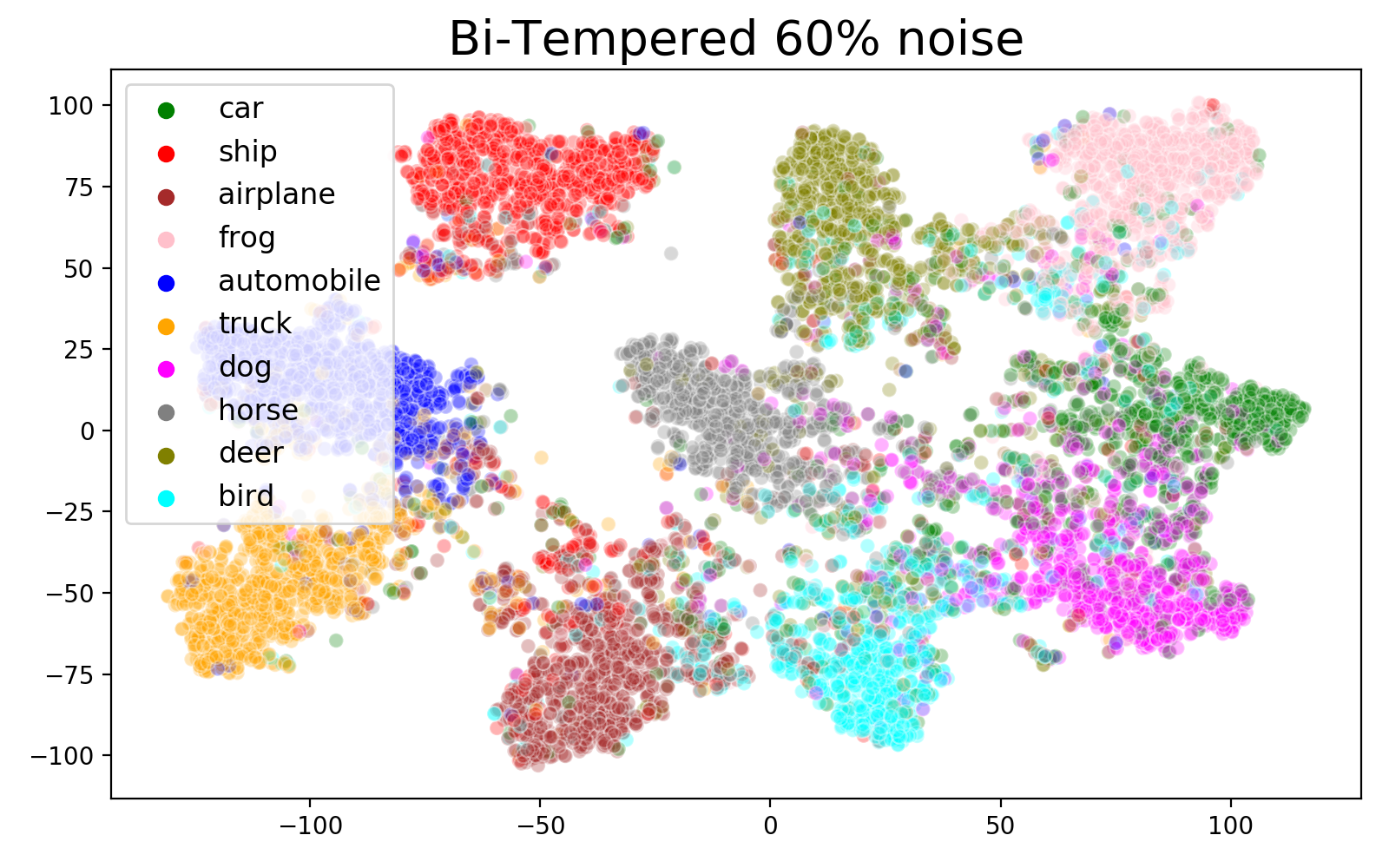}}
	\subfigure[A-Bi-Tem on 60\% noise data.]{
		\label{fig1b} %% label for first subfigure
		\includegraphics[width=0.23\textwidth]{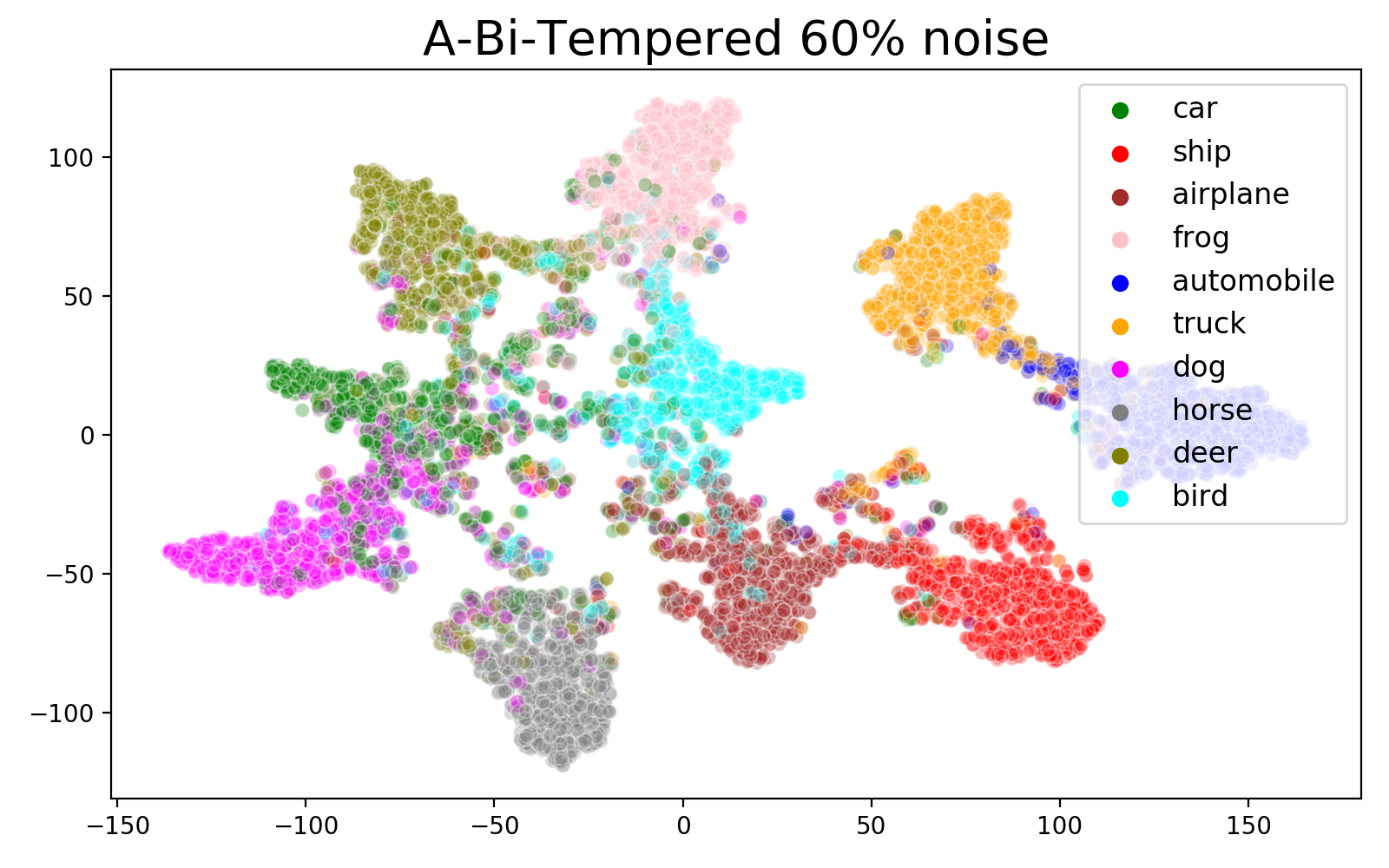}} \\
	\caption{2D representations extracted by A-Bi-Tempered and baselines on CIFAR-10 dataset with 60\% symmetric noisy labels.}\label{figre}
	\vspace{-4mm}
\end{figure}

\begin{figure*}[t]
	\centering
	%\vspace{-1.3cm}
	\subfigcapskip=-1mm
	\subfigure[CIFAR-10 40\% Symmetric Noise]{
		\label{fig1a} %% label for first subfigure
		\includegraphics[width=0.43\textwidth]{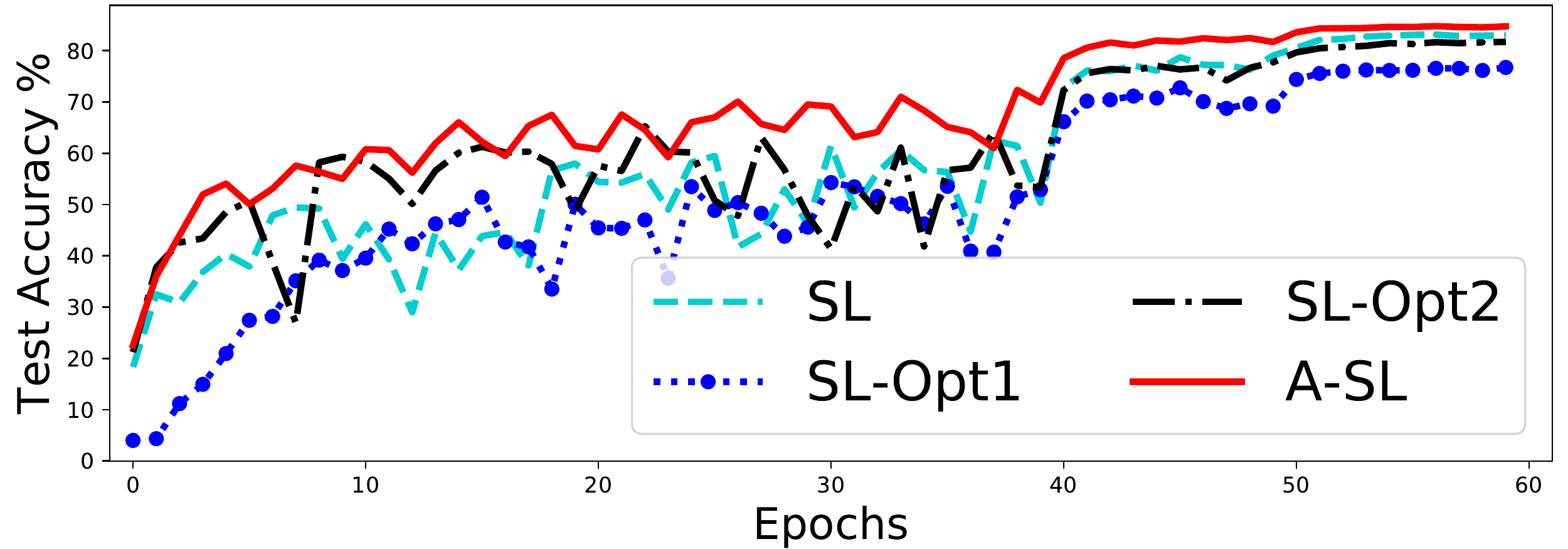}} \ \ \
	\subfigure[CIFAR-10 60\% Symmetric Noise]{
		\label{fig1b} %% label for first subfigure
		\includegraphics[width=0.43\textwidth]{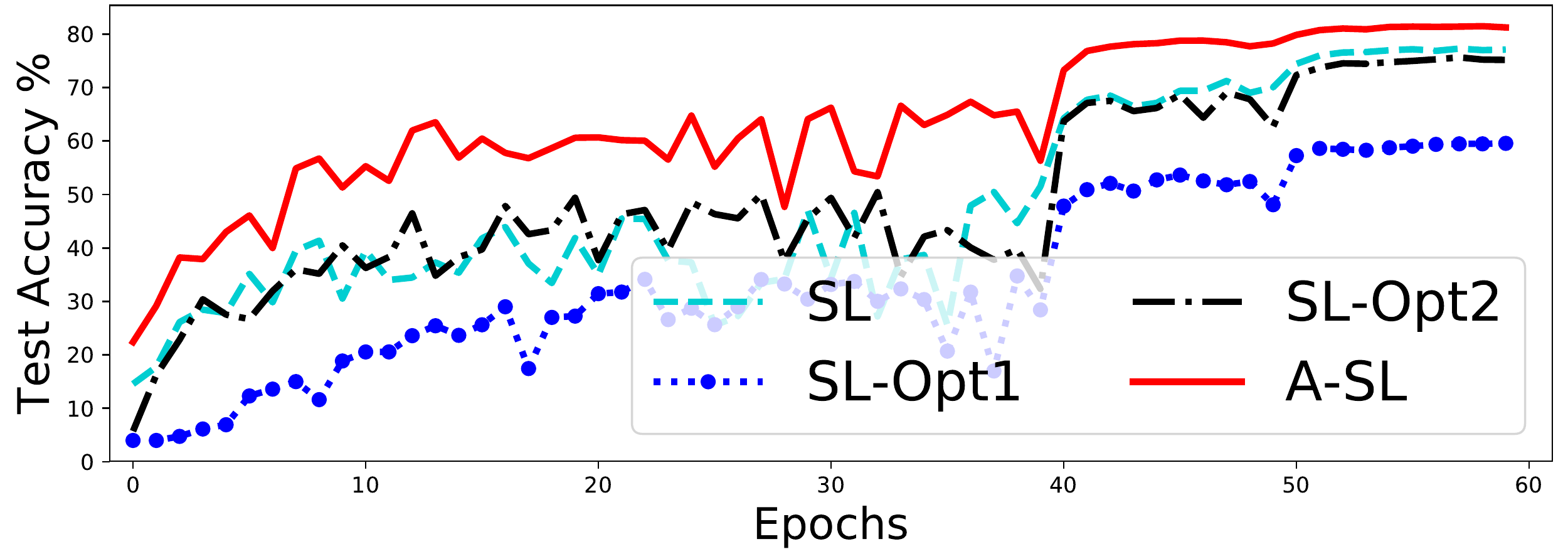}} \\ \vspace{-3mm}
	\subfigure[CIFAR-100 40\% Symmetric Noise]{
		\label{fig1a} %% label for first subfigure
		\includegraphics[width=0.43\textwidth]{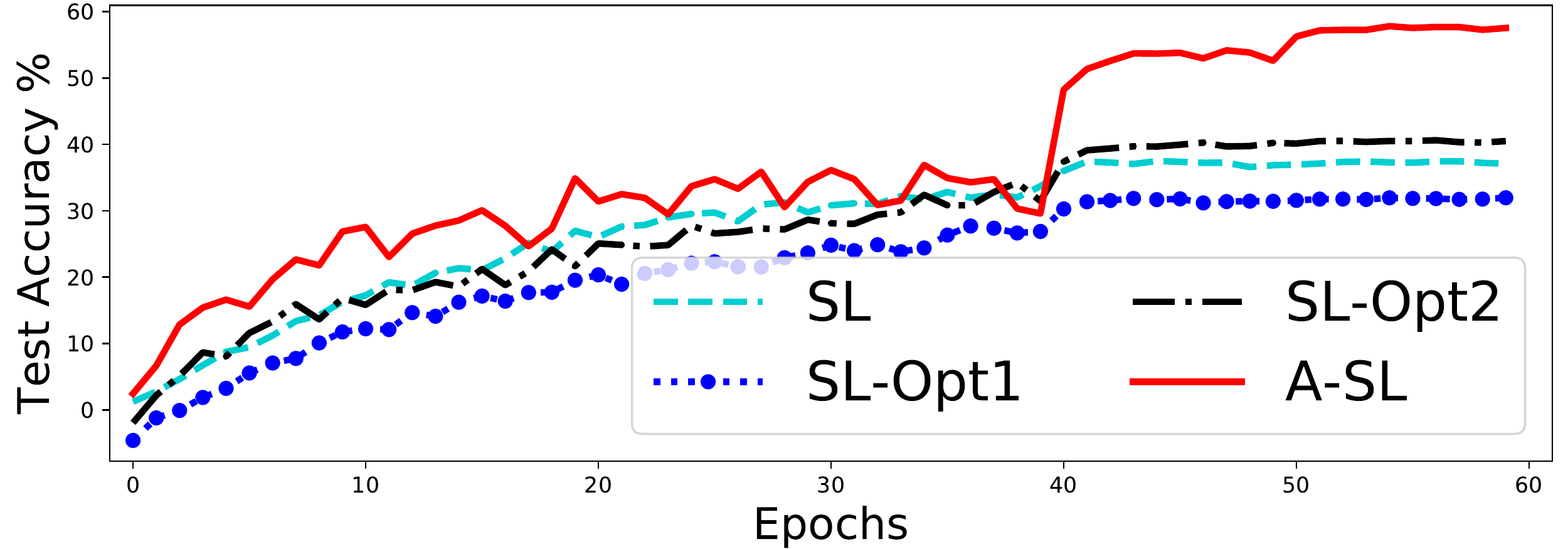}} \ \ \
	\subfigure[CIFAR-100 60\% Symmetric Noise]{
		\label{fig1b} %% label for first subfigure
		\includegraphics[width=0.43\textwidth]{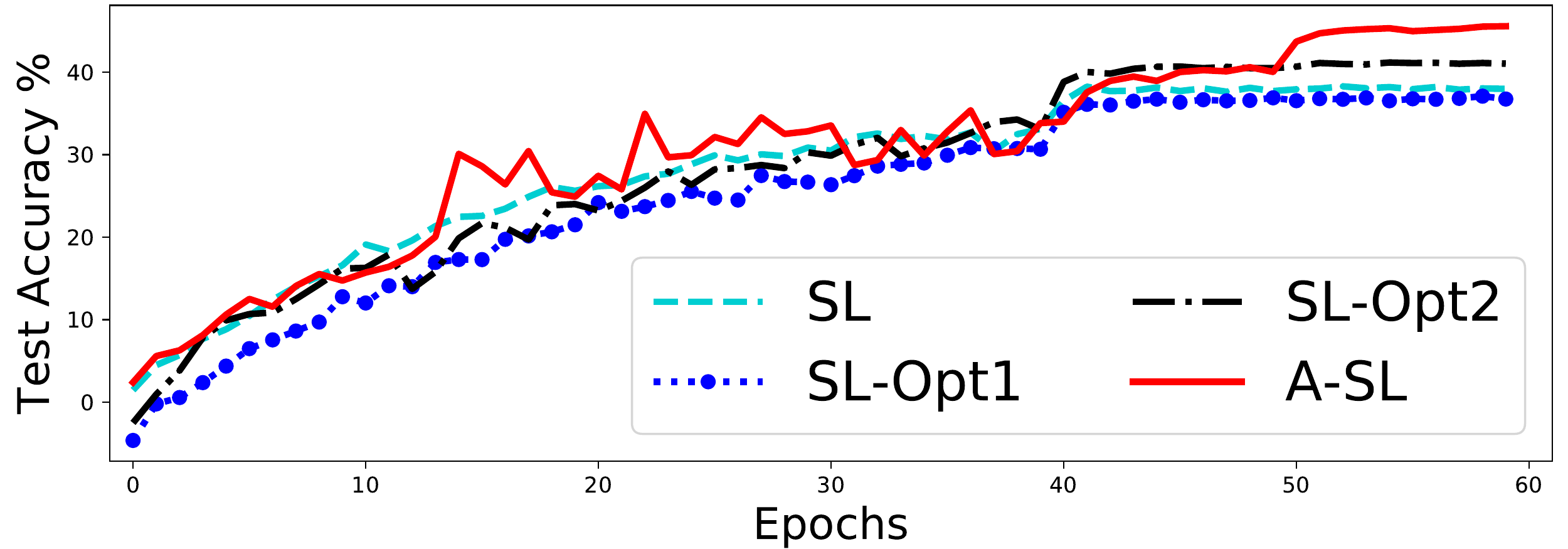}} \\ \vspace{-1mm}
	\caption{Test accuracy vs. number of epochs of A-SL and other comparison methods, SL, SL-Opt1, SL-Opt2 with different noise amounts on CIFAR-10 and CIFAR-100 datasets under symmetric noise.}\label{figSL}
	\vspace{-4mm}
\end{figure*}

\textbf{Reweighting mechanism visualization.} To better understand why our algorithm  contributes to learn more robust models during training, we plot the weight distribution variations of clean and noisy training samples during the learning process of A-PolySoft in Fig.\ref{fig2}. To better visualize this point, we also compare such weight distribution of Meta-Weight-Net. It can be seen the weights extracted by A-PolySoft clearly distinguish clean and noisy samples, much more evident than those obtained by Meta-Weight-Net. Specifically, through the iteration of our algorithm, the weights of clean samples are with larger values gradually than those on noisy ones (most approximates 0), and thus the negative influence of these noisy labels tends to be effectively reduced. This clearly explains why our algorithm is able to consistently outperform the Mete-Weight-Net.

\textbf{Representation Demonstration.} We further investigate the representations learned by ARL algorithm compared to other baselines. We extract the high-dimensional representations of data at the second last dense layer of the learned classifiers by different methods, and then project them to a 2D embedding by using t-SNE \cite{maaten2008visualizing}.
As shown in Fig.\ref{figre}, it is evident that the representations learned by A-Bi-Tempered algorithm (other methods are presented in the supplemental file) are obviously better than that by CE and Bi-Tempered with more separated and clearly bounded clusters.

\vspace{-1mm}
\subsection{Ablation Study}\vspace{-1mm}
ARL algorithm tries to mutually ameliorate robust loss hyperparameters and net parameters in an iterative manner.  An important problem is whether or not the mutual amelioration process helps explore better generalization solution. To clarify this, Fig.\ref{figSL} compares four strategies as: 1) SL: conventional SL method using the hyperparameter optimally tuned by cross-validation; 2) SL-Opt1: conventional SL method using the hyperparameter learned by A-SL at the last step; 3) SL-Opt2: run conventional SL in each step of A-SL by using the hyperparameter obtained by the latter in its current step as its initialization; 4) A-SL: our algorithm. It can be easily observed: 1) SL-Opt1 performs worse than SL, which means the hyperparameter adaptively learned by A-SL is actually not the optimal one for SL, with fixed hyperparameter throughout its iteration.
2) SL-Opt2 outperforms SL, implying the A-SL adaptively finds a proper hyperparameter for its robust loss and simultaneously explores a good initialization net parameter for this loss under its current hyperparameter in a dynamical mutual updating way.
3) ASL outperforms SL-Opt2, showing such adaptive learning manner for both robust loss hyperparameter and net parameters should be a more suitable manner for simultaneously obtaining optimal values for both of them rather than only updating one under the other fixed, even the fixed one could be set possibly optimal. All of these observations inspire such a meta-learning regime might provide a rational learning manner for exploring better generalization solutions for such non-convex robust learning problem.
\vspace{-1mm}
\subsection{Experiments on Real-world Noisy Dataset} \label{real-experiment}\vspace{-1mm}
We then verify the applicability of our algorithm on a real-world large-scale noisy dataset: Clothing1M, which contains 1 million images of clothing from online shopping websites with 14 classes, e.g., T-shirt, Shirt, Knitwear. The labels are generated by the surrounding text of images and are thus extremely noisy. The dataset also provides 50k, 14k, 10k manually refined clean data for training, validation and testing respectively, but we did not use the 50k clean data. We use the validation dataset as the meta dataset.

\textbf{Experimental setup.} Following the previous works\cite{patrini2017making,tanaka2018joint}, we used ResNet-50 pre-trained on ImageNet.  For preprocessing, we resize the image to $256\times256$, crop the center $224\times224$ as input, and perform normalization.  We used SGD with a momentum 0.9, a weight decay $10^{-3}$, an initial learning rate 0.01, and batch size 32. The learning rate of ResNet-50 is divided by 10 after 5 epochs (for a total 10 epochs). We use SGD to optimize hyperparameters with an initial learning rate 0.1, and divided by 10 after 5 epoch.

\textbf{Results.}  The results are summarized in Table \ref{Table4}. The conventional algorithm need to search a proper hyperparameter from a candidate set by cross-validation to obtain a satisfied result, which is often expensive and hard to be reproduced in real-world settings. Our ARL algorithm provides a new way to mutually ameliorate between hyperparameter and network parameters to reduce the barrier of practical implementations. It can be seen that ARL algorithm can consistently improve the performance of the original algorithm, and A-PolySoft obtains the highest performance compared to the baselines.
\vspace{-1mm}
\section{Conclusion} \label{conclusion}\vspace{-1mm}
In this paper, we have proposed an adaptive hyperparameter learning strategy to learn the form of robust loss function directly from data by automatically tuning the hyperparameter. Four STOA robust loss functions are chosen to be integrated into our ARL framework, to verify its validity. Comprehensive experiments have been conducted, and the empirical results show that the propose method can perform superior than conventional hyperparameter setting strategy. The learning fashion of iterative amelioration betwen hyperparameter and network parameter has shown good potential for providing a new thought to explore solutions with better generalization for such highly non-convex robust loss optimization problems.

\newpage

{\small
\bibliographystyle{ieee_fullname}
\bibliography{egbib}

\begin{thebibliography}{10}\itemsep=-1pt

\bibitem{amid2019robust}
Ehsan Amid, Manfred~K Warmuth, Rohan Anil, and Tomer Koren.
\newblock Robust bi-tempered logistic loss based on bregman divergences.
\newblock In {\em NeurIPS}, 2019.

\bibitem{bartlett2006convexity}
Peter~L Bartlett, Michael~I Jordan, and Jon~D McAuliffe.
\newblock Convexity, classification, and risk bounds.
\newblock {\em Journal of the American Statistical Association},
  101(473):138--156, 2006.

\bibitem{bergstra2012random}
James Bergstra and Yoshua Bengio.
\newblock Random search for hyper-parameter optimization.
\newblock {\em JMLR}, 2012.

\bibitem{bi2014learning}
Wei Bi, Liwei Wang, James~T Kwok, and Zhuowen Tu.
\newblock Learning to predict from crowdsourced data.
\newblock In {\em UAI}, 2014.

\bibitem{brooks2011support}
J~Paul Brooks.
\newblock Support vector machines with the ramp loss and the hard margin loss.
\newblock {\em Operations research}, 59(2):467--479, 2011.

\bibitem{chang2017active}
Haw-Shiuan Chang, Erik Learned-Miller, and Andrew McCallum.
\newblock Active bias: Training more accurate neural networks by emphasizing
  high variance samples.
\newblock In {\em NeurIPS}, 2017.

\bibitem{finn2017model}
Chelsea Finn, Pieter Abbeel, and Sergey Levine.
\newblock Model-agnostic meta-learning for fast adaptation of deep networks.
\newblock In {\em ICML}, 2017.

\bibitem{franceschi2017forward}
Luca Franceschi, Michele Donini, Paolo Frasconi, and Massimiliano Pontil.
\newblock Forward and reverse gradient-based hyperparameter optimization.
\newblock In {\em ICML}, 2017.

\bibitem{franceschi2018bilevel}
Luca Franceschi, Paolo Frasconi, Saverio Salzo, Riccardo Grazzi, and
  Massimiliano Pontil.
\newblock Bilevel programming for hyperparameter optimization and
  meta-learning.
\newblock In {\em ICML}, 2018.

\bibitem{ghosh2017robust}
Aritra Ghosh, Himanshu Kumar, and PS Sastry.
\newblock Robust loss functions under label noise for deep neural networks.
\newblock In {\em AAAI}, 2017.

\bibitem{ghosh2015making}
Aritra Ghosh, Naresh Manwani, and PS Sastry.
\newblock Making risk minimization tolerant to label noise.
\newblock {\em Neurocomputing}, 160:93--107, 2015.

\bibitem{goldberger2016training}
Jacob Goldberger and Ehud Ben-Reuven.
\newblock Training deep neural-networks using a noise adaptation layer.
\newblock In {\em ICLR}, 2017.

\bibitem{gong2018decomposition}
Maoguo Gong, Hao Li, Deyu Meng, Qiguang Miao, and Jia Liu.
\newblock Decomposition-based evolutionary multiobjective optimization to
  self-paced learning.
\newblock {\em IEEE Transactions on Evolutionary Computation}, 23(2):288--302,
  2018.

\bibitem{grabocka2019learning}
Josif Grabocka, Randolf Scholz, and Lars Schmidt-Thieme.
\newblock Learning surrogate losses.
\newblock {\em arXiv preprint arXiv:1905.10108}, 2019.

\bibitem{he2016deep}
Kaiming He, Xiangyu Zhang, Shaoqing Ren, and Jian Sun.
\newblock Deep residual learning for image recognition.
\newblock In {\em CVPR}, 2016.

\bibitem{hendrycks2018using}
Dan Hendrycks, Mantas Mazeika, Duncan Wilson, and Kevin Gimpel.
\newblock Using trusted data to train deep networks on labels corrupted by
  severe noise.
\newblock In {\em NeurIPS}, 2018.

\bibitem{huang2019addressing}
Chen Huang, Shuangfei Zhai, Walter Talbott, Miguel~Bautista Martin, Shih-Yu
  Sun, Carlos Guestrin, and Josh Susskind.
\newblock Addressing the loss-metric mismatch with adaptive loss alignment.
\newblock In {\em ICML}, 2019.

\bibitem{huber2011robust}
Peter~J Huber.
\newblock {\em Robust statistics}.
\newblock Springer, 2011.

\bibitem{jiang2014easy}
Lu Jiang, Deyu Meng, Teruko Mitamura, and Alexander~G Hauptmann.
\newblock Easy samples first: Self-paced reranking for zero-example multimedia
  search.
\newblock In {\em Proceedings of the 22nd ACM international conference on
  Multimedia}, pages 547--556, 2014.

\bibitem{jiang2018mentornet}
Lu Jiang, Zhengyuan Zhou, Thomas Leung, Li-Jia Li, and Li Fei-Fei.
\newblock Mentornet: Learning data-driven curriculum for very deep neural
  networks on corrupted labels.
\newblock In {\em ICML}, 2018.

\bibitem{krizhevsky2009learning}
Alex Krizhevsky.
\newblock Learning multiple layers of features from tiny images.
\newblock Technical report, 2009.

\bibitem{krizhevsky2012imagenet}
Alex Krizhevsky, Ilya Sutskever, and Geoffrey~E Hinton.
\newblock Imagenet classification with deep convolutional neural networks.
\newblock In {\em NeurIPS}, 2012.

\bibitem{kumar2010self}
M~Pawan Kumar, Benjamin Packer, and Daphne Koller.
\newblock Self-paced learning for latent variable models.
\newblock In {\em NeurIPS}, 2010.

\bibitem{lee2018cleannet}
Kuang-Huei Lee, Xiaodong He, Lei Zhang, and Linjun Yang.
\newblock Cleannet: Transfer learning for scalable image classifier training
  with label noise.
\newblock In {\em CVPR}, 2018.

\bibitem{li2017learning}
Yuncheng Li, Jianchao Yang, Yale Song, Liangliang Cao, Jiebo Luo, and Li-Jia
  Li.
\newblock Learning from noisy labels with distillation.
\newblock In {\em ICCV}, 2017.

\bibitem{liang2016learning}
Junwei Liang, Lu Jiang, Deyu Meng, and Alexander~G Hauptmann.
\newblock Learning to detect concepts from webly-labeled video data.
\newblock In {\em IJCAI}, 2016.

\bibitem{liu2014robust}
Anqi Liu and Brian Ziebart.
\newblock Robust classification under sample selection bias.
\newblock In {\em NeurIPS}, 2014.

\bibitem{lyu2019curriculum}
Yueming Lyu and Ivor~W Tsang.
\newblock Curriculum loss: Robust learning and generalization against label
  corruption.
\newblock {\em arXiv preprint arXiv:1905.10045}, 2019.

\bibitem{maaten2008visualizing}
Laurens van~der Maaten and Geoffrey Hinton.
\newblock Visualizing data using t-sne.
\newblock {\em Journal of machine learning research}, 9:2579--2605, 2008.

\bibitem{maclaurin2015gradient}
Dougal Maclaurin, David Duvenaud, and Ryan Adams.
\newblock Gradient-based hyperparameter optimization through reversible
  learning.
\newblock In {\em ICML}, 2015.

\bibitem{manwani2013noise}
Naresh Manwani and PS Sastry.
\newblock Noise tolerance under risk minimization.
\newblock {\em IEEE transactions on cybernetics}, 43(3):1146--1151, 2013.

\bibitem{masnadi2009design}
Hamed Masnadi-Shirazi and Nuno Vasconcelos.
\newblock On the design of loss functions for classification: theory,
  robustness to outliers, and savageboost.
\newblock In {\em NeurIPS}, 2009.

\bibitem{meng2017theoretical}
Deyu Meng, Qian Zhao, and Lu Jiang.
\newblock A theoretical understanding of self-paced learning.
\newblock {\em Information Sciences}, 414:319--328, 2017.

\bibitem{natarajan2013learning}
Nagarajan Natarajan, Inderjit~S Dhillon, Pradeep~K Ravikumar, and Ambuj Tewari.
\newblock Learning with noisy labels.
\newblock In {\em NeurIPS}, 2013.

\bibitem{nock2009efficient}
Richard Nock and Frank Nielsen.
\newblock On the efficient minimization of classification calibrated
  surrogates.
\newblock In {\em NeurIPS}, 2009.

\bibitem{paszke2017automatic}
Adam Paszke, Sam Gross, Soumith Chintala, Gregory Chanan, Edward Yang, Zachary
  DeVito, Zeming Lin, Alban Desmaison, Luca Antiga, and Adam Lerer.
\newblock Automatic differentiation in pytorch.
\newblock In {\em NIPS Workshop}, 2017.

\bibitem{patrini2017making}
Giorgio Patrini, Alessandro Rozza, Aditya Krishna~Menon, Richard Nock, and
  Lizhen Qu.
\newblock Making deep neural networks robust to label noise: A loss correction
  approach.
\newblock In {\em CVPR}, 2017.

\bibitem{pedregosa2016hyperparameter}
Fabian Pedregosa.
\newblock Hyperparameter optimization with approximate gradient.
\newblock In {\em ICML}, 2016.

\bibitem{reed2014training}
Scott Reed, Honglak Lee, Dragomir Anguelov, Christian Szegedy, Dumitru Erhan,
  and Andrew Rabinovich.
\newblock Training deep neural networks on noisy labels with bootstrapping.
\newblock In {\em ICLR}, 2015.

\bibitem{ren2018learning}
Mengye Ren, Wenyuan Zeng, Bin Yang, and Raquel Urtasun.
\newblock Learning to reweight examples for robust deep learning.
\newblock In {\em ICML}, 2018.

\bibitem{schmidhuber1992learning}
J{\"u}rgen Schmidhuber.
\newblock Learning to control fast-weight memories: An alternative to dynamic
  recurrent networks.
\newblock {\em Neural Computation}, 4(1):131--139, 1992.

\bibitem{shu2019meta}
Jun Shu, Qi Xie, Lixuan Yi, Qian Zhao, Sanping Zhou, Zongben Xu, and Deyu Meng.
\newblock Meta-weight-net: Learning an explicit mapping for sample weighting.
\newblock In {\em NeurIPS}, 2019.

\bibitem{shu2018small}
Jun Shu, Zongben Xu, and Deyu Meng.
\newblock Small sample learning in big data era.
\newblock {\em arXiv preprint arXiv:1808.04572}, 2018.

\bibitem{snoek2012practical}
Jasper Snoek, Hugo Larochelle, and Ryan~P Adams.
\newblock Practical bayesian optimization of machine learning algorithms.
\newblock In {\em NeurIPS}, 2012.

\bibitem{sukhbaatar2014training}
Sainbayar Sukhbaatar, Joan Bruna, Manohar Paluri, Lubomir Bourdev, and Rob
  Fergus.
\newblock Training convolutional networks with noisy labels.
\newblock In {\em ICLR Workshops}, 2015.

\bibitem{swersky2013multi}
Kevin Swersky, Jasper Snoek, and Ryan~P Adams.
\newblock Multi-task bayesian optimization.
\newblock In {\em NeurIPS}, 2013.

\bibitem{tanaka2018joint}
Daiki Tanaka, Daiki Ikami, Toshihiko Yamasaki, and Kiyoharu Aizawa.
\newblock Joint optimization framework for learning with noisy labels.
\newblock In {\em CVPR}, 2018.

\bibitem{thrun2012learning}
Sebastian Thrun and Lorien Pratt.
\newblock {\em Learning to learn}.
\newblock Springer, 1998.

\bibitem{vahdat2017toward}
Arash Vahdat.
\newblock Toward robustness against label noise in training deep discriminative
  neural networks.
\newblock In {\em NeurIPS}, 2017.

\bibitem{van2015learning}
Brendan Van~Rooyen, Aditya Menon, and Robert~C Williamson.
\newblock Learning with symmetric label noise: The importance of being
  unhinged.
\newblock In {\em NeurIPS}, 2015.

\bibitem{veit2017learning}
Andreas Veit, Neil Alldrin, Gal Chechik, Ivan Krasin, Abhinav Gupta, and Serge
  Belongie.
\newblock Learning from noisy large-scale datasets with minimal supervision.
\newblock In {\em CVPR}, 2017.

\bibitem{wang2017robust}
Yixin Wang, Alp Kucukelbir, and David~M Blei.
\newblock Robust probabilistic modeling with bayesian data reweighting.
\newblock In {\em ICML}, 2017.

\bibitem{wang2019symmetric}
Yisen Wang, Xingjun Ma, Zaiyi Chen, Yuan Luo, Jinfeng Yi, and James Bailey.
\newblock Symmetric cross entropy for robust learning with noisy labels.
\newblock In {\em ICCV}, 2019.

\bibitem{wu2018learning}
Lijun Wu, Fei Tian, Yingce Xia, Yang Fan, Tao Qin, Lai Jian-Huang, and Tie-Yan
  Liu.
\newblock Learning to teach with dynamic loss functions.
\newblock In {\em NeurIPS}, 2018.

\bibitem{xiao2015learning}
Tong Xiao, Tian Xia, Yi Yang, Chang Huang, and Xiaogang Wang.
\newblock Learning from massive noisy labeled data for image classification.
\newblock In {\em CVPR}, 2015.

\bibitem{xu2018autoloss}
Haowen Xu, Hao Zhang, Zhiting Hu, Xiaodan Liang, Ruslan Salakhutdinov, and Eric
  Xing.
\newblock Autoloss: Learning discrete schedules for alternate optimization.
\newblock In {\em ICLR}, 2019.

\bibitem{xu2019l_dmi}
Yilun Xu, Peng Cao, Yuqing Kong, and Yizhou Wang.
\newblock L\_dmi: An information-theoretic noise-robust loss function.
\newblock In {\em NeurIPS}, 2019.

\bibitem{zhang2016understanding}
Chiyuan Zhang, Samy Bengio, Moritz Hardt, Benjamin Recht, and Oriol Vinyals.
\newblock Understanding deep learning requires rethinking generalization.
\newblock In {\em ICLR}, 2017.

\bibitem{zhang2018generalized}
Zhilu Zhang and Mert Sabuncu.
\newblock Generalized cross entropy loss for training deep neural networks with
  noisy labels.
\newblock In {\em NeurIPS}, 2018.

\bibitem{zhao2015self}
Qian Zhao, Deyu Meng, Lu Jiang, Qi Xie, Zongben Xu, and Alexander~G Hauptmann.
\newblock Self-paced learning for matrix factorization.
\newblock In {\em AAAI}, 2015.

\bibitem{zhuang2017attend}
Bohan Zhuang, Lingqiao Liu, Yao Li, Chunhua Shen, and Ian Reid.
\newblock Attend in groups: a weakly-supervised deep learning framework for
  learning from web data.
\newblock In {\em CVPR}, 2017.

\end{thebibliography}
}

\end{document}